\documentclass[10pt,twocolumn,letterpaper]{article}

\usepackage[letterpaper,margin=0.72in,columnsep=0.24in]{geometry}
\usepackage{times}
\usepackage{graphicx}
\usepackage{amsmath}
\usepackage{amssymb}
\usepackage{amsthm}
\usepackage{booktabs}
\usepackage{array}
\usepackage{multirow}
\usepackage{microtype}
\usepackage{placeins}
\usepackage{caption}
\usepackage{xcolor}
\usepackage[numbers,sort&compress]{natbib}

\definecolor{linkblue}{rgb}{0.10,0.35,0.62}
\usepackage[breaklinks,colorlinks,allcolors=linkblue]{hyperref}

\newtheorem{theorem}{Theorem}
\newtheorem{proposition}[theorem]{Proposition}
\newtheorem{corollary}[theorem]{Corollary}

\title{Beyond Prediction Gain: LASR as a Metric for Action-Grounded Latents in World Models}
\author{
Yu Zhou$^{1}$, Changsong Lei$^{1}$, Yixiang Wei$^{1}$, Kairui Wen$^{1}$,\\
Huankun Sheng$^{1}$, Ming Li$^{1}$, Yeying Fan$^{1}$, Yu-Hui Wen$^{2}$, Yong-Jin Liu$^{1,*}$\\[0.45em]
$^{1}$Tsinghua University \qquad $^{2}$Beijing Jiaotong University\\
$^{*}$Corresponding author\\[0.3em]
\small\texttt{\{zhou-y24,leics23,yx-wei22,wkr21\}@mails.tsinghua.edu.cn}\\
\small\texttt{\{shenghk,mingli\_thu,fanyeying,liuyongjin\}@tsinghua.edu.cn}\\
\small\texttt{yhwen1@bjtu.edu.cn}
}

\hypersetup{
  pdftitle={Beyond Prediction Gain: LASR as a Metric for Action-Grounded Latents in World Models},
  pdfauthor={Yu Zhou, Changsong Lei, Yixiang Wei, Kairui Wen, Huankun Sheng, Ming Li, Yeying Fan, Yu-Hui Wen, Yong-Jin Liu}
}
\date{}

\begin{document}

\maketitle

\begin{abstract}
World models that support planning, control, and interpretable reasoning require latent representations that genuinely encode action. Yet the large-scale video data from which such models are learned overwhelmingly lack action annotations. Latent-action models infer action-like variables from observation sequences, enabling action-aware learning from unlabeled videos or trajectories. A fundamental concern, however, remains: a latent that improves future prediction may simply encode the current state, episode progress, task identity, or other predictive shortcuts. Existing evaluation practices---reconstruction loss, rollout fidelity, action recovery, or downstream performance---cannot isolate this distinction. We introduce the Latent Action Support Ratio (LASR), a dimensionless metric that measures, under offline evaluation on held-out data, whether a learned latent provides predictive support for action-related targets beyond the current state and non-action alternatives. LASR provides two complementary summaries: GeoLASR captures typical support, while WorstLASR identifies the weakest case. We instantiate the LASR benchmark with recorded actions and corresponding physical state changes as targets, episode-held-out splits, one predictor type for all comparisons, and positive and negative controls. Experiments on recorded robot manipulation data and a matched multi-task panel demonstrate that LASR detects known action information, rejects representative non-action shortcuts, and enables graded comparisons among imperfect latent representations. We further use LASR to train and audit a lightweight residual adapter, obtaining descriptively better point estimates whose uncertainty is assessed with paired comparisons. LASR thus provides a principled offline protocol for measuring action-related predictive support in world-model latents.
\end{abstract}

\section{Introduction}
World models intended for planning and control need to capture how an agent's actions affect future states \citep{saanum2024simplifying,hafner2019learning}. However, the large-scale video data used to train such models rarely include synchronized low-level action annotations, while robot trajectories with such labels are costly to collect and limited in task and embodiment coverage \citep{ye2025latent,bu2025univla,zhang2026dila}. This mismatch poses a central challenge for embodied learning: a world model can make future predictions, whereas planning and control require a representation of how an agent's actions change the future. 

Latent-action models seek to bridge this gap by inferring compact action-like variables from consecutive observations, thereby enabling action-conditioned learning from unlabeled video~\citep{ye2025latent,bu2025univla,zhang2026dila}. These inferred variables are treated as latent actions and are intended to serve as proxies for the unobserved physical actions that connect successive observations. Existing evaluations of latent-action models, however, raise a basic question \citep{zhang2026latent}: when a latent improves prediction, does it contribute action-related information, or does it merely preserve predictive structure already present in the logged trajectory?

\begin{figure*}[t]
\centering
\textbf{(a) State and task-phase predictability}\par
\includegraphics[width=\textwidth]{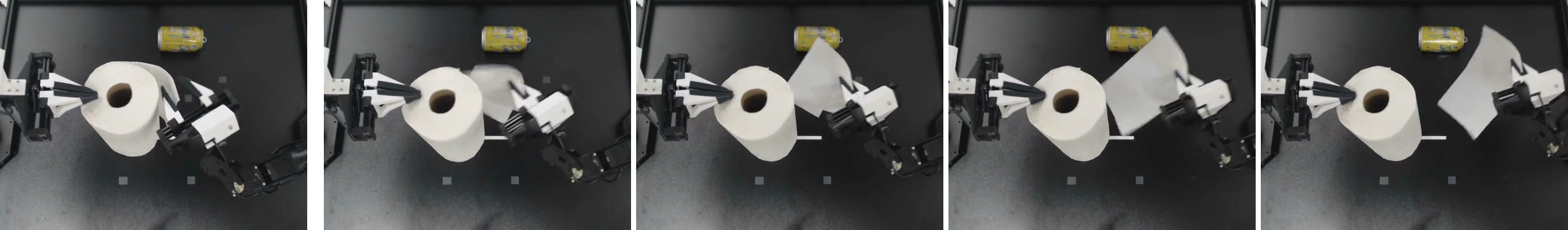}
\vspace{2pt}
\textbf{(b) Scene and episode identity}\par
\includegraphics[width=\textwidth]{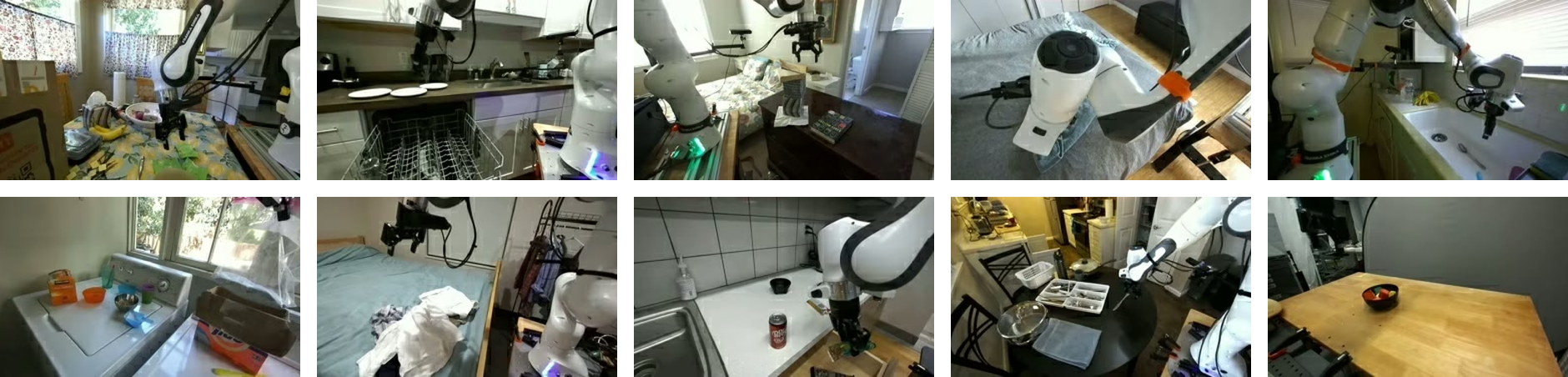}
\caption{Two shortcuts in offline latent-action evaluation. (a) A trajectory from the ALOHA static-battery release illustrates how the current state and task phase constrain the next motion \citep{zhao2023learning}. (b) DROID-100 observations from distinct scenes, viewpoints, and object layouts illustrate how scene and episode identity may help predict typical transitions without isolating action-specific information \citep{khazatsky2024droid}.}
\label{fig:offline-shortcuts}
\end{figure*}

Distinguishing action-related information from predictable trajectory structure is especially challenging in offline robot datasets, where the current state alone may strongly constrain the demonstrator's next action. For instance, Figure~\ref{fig:offline-shortcuts}(a) shows a trajectory from the ALOHA static-battery release in which the grippers and manipulated material are already in contact \citep{zhao2023learning}. The frame therefore reveals both the contact state and the task phase, making the subsequent motion more predictable even without a latent action. A latent that preserves these state cues may improve future prediction without contributing additional action-related information beyond the current state. More generally, the scene and episode cues visible in Figure~\ref{fig:offline-shortcuts}(b) may help predict typical transitions without representing the agent's action, and episode progress and temporal regularities can provide similar non-action shortcuts \citep{de2019causal,nikulin2025latent}.

A second confound arises from the evaluation model itself. A predictor that receives both the current state and a latent may have greater effective capacity than a predictor that receives the state alone, so a lower prediction loss need not reflect action-related information in the latent \citep{hewitt2019designing}. Consequently, transition-prediction loss, reconstruction quality, action-prediction accuracy, and downstream policy success cannot, in isolation, determine whether improved performance reflects action-related information rather than state predictability, non-action shortcuts, or differences in evaluation capacity.

\begin{table*}[t]
\centering
\small
\setlength{\tabcolsep}{5pt}
\begin{tabular}{p{0.22\textwidth}p{0.70\textwidth}}
\hline
\textbf{Method} & \textbf{Reported evaluation metrics} \\
\hline
LAPA~\citep{ye2025latent} &
Task success and task score; qualitative latent-token and rollout analyses \\
UniVLA~\citep{bu2025univla} &
Manipulation success, step-wise real-robot score, and navigation oracle success \\
MVP-LAM~\citep{lee2026mvp} &
Estimated $I(Z;A)$, linear action-prediction NMSE, DINOv2 feature-space reconstruction MSE under viewpoint perturbation, and task success \\
DiLA~\citep{zhang2026dila} &
Rollout SSIM/LPIPS, cycle-transfer LPIPS, linear robot-action prediction MSE, and visual-planning success \\
LaWAM~\citep{chen2026lawam} &
LIBERO, RoboTwin, and real-robot success, action-chunk latency, and component-ablation success \\
LASR (ours) &
LASR scores by target, GeoLASR, WorstLASR, and paired uncertainty \\
\hline
\end{tabular}
\caption{Representative metrics used to evaluate latent-action and world-action models.}
\label{tab:prior-evaluation-styles}
\end{table*}

Table~\ref{tab:prior-evaluation-styles} shows that prior work evaluates latent-action models using heterogeneous outcomes, including policy success, latent-action correlation, reconstruction fidelity, and planning performance. Because these quantities measure different objects on different scales, their values do not define a common measure of latent quality. We therefore formulate latent-action evaluation as a {\it metric-design problem}: an evaluation metric must first specify what it means for a learned latent to function as an action. We adopt a functional definition in which a learned latent functions as an action to the extent that it contributes action-related information about an evaluation target beyond what is available from the current state and from plausible non-action shortcuts. We call this property \emph{latent action support}. We use the recorded physical action and one-step physical state change as targets, and compare the latent with the current state and non-action variables such as episode progress and episode identity.

We introduce the Latent Action Support Ratio (LASR), a dimensionless metric for measuring whether a learned latent provides predictive support for action-related targets beyond what the current state and a plausible non-action control can explain. A lower LASR score indicates stronger support, with one serving as the operational boundary. Across targets and data splits, LASR offers two complementary summaries: GeoLASR summarizes typical support, whereas WorstLASR exposes the weakest case. Known action and non-action examples check whether the metric behaves as intended, while our analysis explains why the comparison captures information contributed by the latent.

Building on the LASR metric, we also construct the LASR benchmark, which provides a common setting for applying LASR to different representations. We further use LASR's relative comparison to train a small residual adapter and test whether it makes action-related information in a frozen latent easier to use. We compare the adapted representation with the original latent and several compression and training alternatives, and use paired uncertainty to assess the reliability of the differences. Additional matched multi-task, probe/control, physical-replay, and intervention audits test whether the main interpretation survives broader confounds without changing the metric definition.

We make the following three contributions:
\begin{itemize}
\item We introduce LASR, a dimensionless metric for quantifying latent action support. Building on LASR, we define GeoLASR and WorstLASR as complementary aggregate summaries of typical and worst-case support, respectively, and construct the LASR benchmark for consistent comparisons across representations.
\item We provide a theoretical characterization of latent-action information through a conditional Bayes-risk identity, oracle retention, and an information boundary for deterministic post-hoc adapters.
\item Experiments on robot manipulation data show that LASR detects known action information, rejects representative non-action shortcuts, and supports graded comparisons among imperfect latent representations. Matched multi-task and intervention audits broaden this validation, while a lightweight residual adapter illustrates a secondary use of the LASR-inspired surrogate under paired uncertainty.
\end{itemize}

\section{Related Work}

\paragraph{Latent actions from video to control.} Action-free pretraining learns control-relevant dynamics without recorded actions, from video-based reinforcement-learning transfer to joint inverse--forward dynamics objectives \citep{seo2022reinforcement,cui2024dynamo}. LAPO converts observation changes into discrete latent controls for policy learning, while Genie uses learned latent actions to make video-trained environments interactive \citep{schmidt2024learning,bruce2024genie}. In robotics, learned action abstractions support assistive control and bridge unlabeled videos to VLA pretraining through pseudo-actions or motion tokens \citep{losey2022learning,ye2025latent,chen2025moto,bu2025univla,chen2025villa}. A controlled study further compares how image- and action-based latent supervision should enter a shared VLA backbone \citep{lin2026pixels}. Parallel world- and action-model work uses latent transition codes for adaptable generation, continuous control, action-centric reconstruction, disentangled dynamics, and efficient action chunking \citep{gao2025adaworld,liang2025clam,lee2026mvp,zhang2026dila,chen2026lawam}. PreLAR additionally uses action-representation pretraining to improve downstream robot learning~\citep{zhang2024prelar}. Collectively, these methods demonstrate utility through downstream control, transfer, reconstruction, rollout fidelity, or planning. Those outcomes, however, neither define a common scale nor isolate what the exported latent contributes beyond the observed state and plausible non-action information.

\paragraph{World-model and latent-action evaluation.} World-model benchmarks assess perceptual quality, planning utility, or closed-loop task success \citep{qin2024worldsimbench,zhang2025world}, but these system-level outcomes jointly reflect the latent, decoder, controller, and policy rather than isolating the latent's conditional contribution. Representation-level studies compare latent spaces through reconstruction, action recovery, planning, and downstream control, while structural tests show that prediction accuracy and state decodability can mask incoherent dynamics \citep{jha2026reconstruction,vafa2024evaluating}. Linear probes test information accessibility, control tasks expose evaluator memorization, and conditional probes measure contributions beyond a specified reference \citep{alain2016understanding,hewitt2019designing,hewitt2021conditional}; however, improvement over a single reference does not show that the latent outperforms both the state route and a plausible non-action control route under a matched predictor family and fitting procedure. Within latent-action learning, prior work studies variance capture, future-state leakage, action-alignment failures, and identifiability, or develops training objectives for particular confounders \citep{zhang2026latent,lee2026latent,nikulin2025latent,lachapelle2025identifiability,garrido2026learning,jiang2026olaf,bu2026laof,klepach2026object,wei2026causally}. LASR instead defines a per-target--fold score from the held-out prediction risks of the latent, state, and validation-selected control routes. The same relative-risk structure motivates the LASR-inspired surrogate used by our LASR-guided adapter, while the reported metric remains distinct from this training surrogate.

\paragraph{Metric and benchmark design.} Work on SSIM and LPIPS replaces raw pixel error with structural or learned-feature comparisons, FID evaluates generated distributions, and BERTScore and CLIPScore define representation-based semantic alignment \citep{wang2004image,zhang2018unreasonable,heusel2017gans,zhang2019bertscore,hessel2021clipscore}. VBench and GenEval further decompose broad generation quality into declared dimensions and controlled cases that expose model-specific failures \citep{huang2024vbench,ghosh2023geneval}. The transferable lesson is methodological: define the evaluation object, validate the score with domain-appropriate evidence and controls, report failure cases, and make the protocol reproducible. LASR applies this metric-design principle to conditional action support; it does not assume that human perceptual agreement is the appropriate validation target for robot latents.

\section{Method}

We introduce LASR to measure the extent of action-related predictive support provided by a learned latent representation within a world model. Specifically, LASR assesses whether the latent improves held-out prediction over what can already be accounted for by the current state and non-action cues.

\begin{figure*}[t]
\centering
\includegraphics[width=\textwidth]{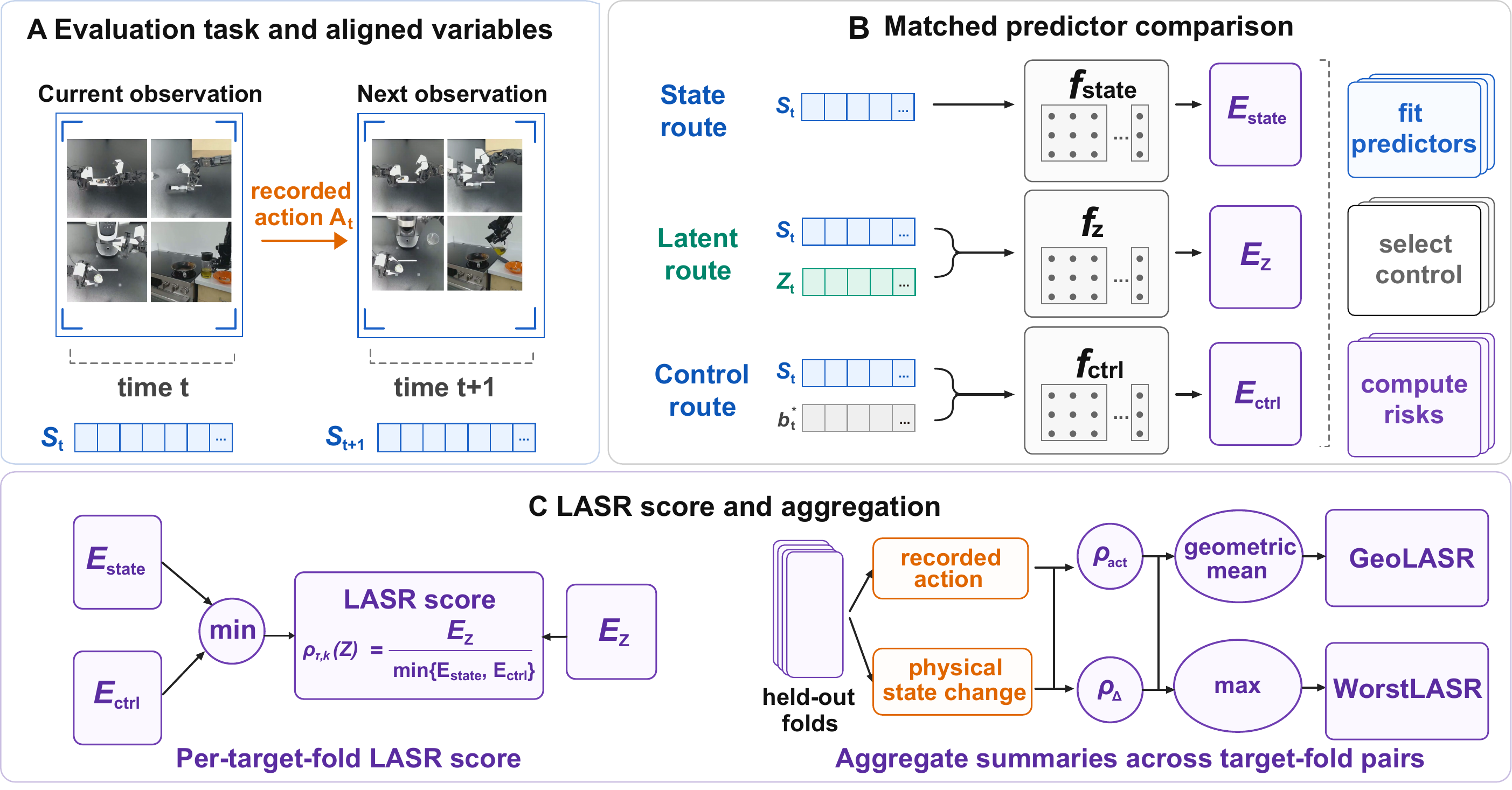}
\caption{Overview of LASR. (A) A logged transition aligns the current state $S_t$, recorded action $A_t$, and next state $S_{t+1}$. (B) For each target--fold pair, separately fitted state, latent, and validation-selected control routes yield held-out risks $E_{\mathrm{state}}$, $E_Z$, and $E_{\mathrm{ctrl}}$. (C) LASR divides $E_Z$ by the smaller of the two reference risks. A score below one means that the latent route outperforms both references; GeoLASR and WorstLASR are the geometric mean and maximum across target--fold scores, respectively. Lower is better.}
\label{fig:lasr-pipeline}
\end{figure*}

The assessment uses three matched predictor routes---state, latent, and control---to provide a controlled comparison. LASR produces a score for each target--fold pair. GeoLASR and WorstLASR then aggregate the per-pair scores to summarize typical and worst-case action support, respectively. We first define the evaluation task and formalize the matched comparison and score, then present the theoretical interpretation, benchmark instantiation, and representation-refinement procedure. Figure~\ref{fig:lasr-pipeline} provides an overview of the evaluation task, matched predictor comparison, and LASR scoring and aggregation.

\subsection{Evaluation Task}

Given a latent representation $Z_t$ aligned with a logged transition, we evaluate whether it provides predictive support for an action-related target $Y_\tau$ beyond the current state $S_t$ and a candidate non-action variable $b$. Here, $S_t$ denotes the synchronized proprioceptive or simulator state used for audit, rather than the RGB observation from which an image-based latent may be extracted. In our LASR benchmark on ALOHA~\citep{zhao2023learning}, we instantiate \(Y_\tau\) with two complementary targets to capture different aspects of action: the recorded physical action $Y_{\mathrm{act}}=A_t$ and the observed one-step physical-state difference $Y_{\Delta}=S_{t+1}-S_t$. We evaluate this on held-out episodes by comparing prediction from $(S_t,Z_t)$ with prediction from $S_t$ alone and from $(S_t,b)$.

\subsection{Matched Predictor Comparisons}

To isolate the predictive support available from the latent representation $Z$ beyond what is already provided by the current state $S$ and plausible non-action variables, we compare three predictor routes. We define a \emph{route} as a specific input configuration together with its separately trained predictor. The state route uses only $S$ and quantifies the baseline predictability of the target $Y_\tau$ from the current state alone; the latent route uses $(S,Z)$ and measures the additional predictive support available from the learned latent representation; and the control route uses $(S,b)$ for a candidate non-action variable $b\in\mathcal C$ and tests whether a non-action variable could produce a similar improvement.

All three routes predict the same target $Y_\tau$ using the same loss function and predictor family; they differ only in their input information. The state and control routes serve as competing reference baselines, while the latent route is the primary target of evaluation. We denote their respective prediction risks by $E_{\mathrm{state}}$, $E_Z$, and $E_{\mathrm{ctrl}}$, where $E_{\mathrm{ctrl}}$ is the held-out risk of the control selected from $\mathcal C$ by validation risk; the exact selection rule is given in Appendix Sec.~A.

\subsection{LASR Score}

The aforementioned prediction risks are not directly comparable across different targets because target scale, dimensionality, and loss scale may differ. To address this, LASR defines a dimensionless score for each target~$\tau$ and held-out fold~$k$:
\begin{equation}
 \rho_{\tau,k}(Z)
 =\frac{E_Z}
 {\min\{E_{\mathrm{state}},E_{\mathrm{ctrl}}\}}.
 \label{eq:target-lasr}
\end{equation}
Here, all three risk terms refer to the same target--fold pair $(\tau,k)$, with the indices suppressed for readability; $E_{\mathrm{ctrl}}$ is the held-out test risk of the control selected on validation episodes.
The score normalizes the latent-route risk by the stronger of the two reference routes, selected by the minimum. Thus, to satisfy the LASR criterion, the latent representation should outperform \emph{both} reference routes, not just the weaker one.

When $\rho_{\tau,k}(Z)<1$, the latent route achieves lower risk than both references, satisfying the LASR criterion for action-grounded predictive support. When $\rho_{\tau,k}(Z)\geq 1$, at least one reference route performs as well as or better than the latent route. However, this does not prove that the latent representation lacks action-related information. LASR is inherently relative---its outcome depends on the chosen target, predictor family, non-action variables, and available data.

Because each LASR score is a relative-risk ratio, a common positive rescaling of all three held-out risks leaves the score unchanged. Formal treatments of the threshold interpretation, aggregation, scale invariance, and control monotonicity are provided in Appendix Secs.~A and B.1.

\paragraph{Held-out risk estimation.} In practice, route predictors are fitted on training episodes, the non-action variable is selected on validation episodes, and the three risks are estimated on held-out test episodes. The population identities below motivate the comparison, but empirical LASR remains relative to the finite predictor and control families declared by the benchmark. Full details of fold construction, control selection, predictor fitting, and validity checks are provided in Appendix Secs.~C and H.1.

\subsection{GeoLASR and WorstLASR}

We fix a common valid set $\Omega$ for all representations. Shared invalid reference pairs are removed for every representation, whereas representation-specific invalidity does not change $\Omega$. LASR offers two aggregate summaries:
\begin{align}
 \operatorname{GeoLASR}(Z;\Omega)
 &=\left(\prod_{(\tau,k)\in\Omega}\rho_{\tau,k}(Z)\right)^{1/|\Omega|},
 \label{eq:geo-lasr}\\
 \operatorname{WorstLASR}(Z;\Omega)
 &=\max_{(\tau,k)\in\Omega}\rho_{\tau,k}(Z).
 \label{eq:worst-lasr}
\end{align}
Lower is better for both summaries. GeoLASR below one indicates a favorable geometric-average ratio but may include individual failures, whereas WorstLASR is below one if and only if every target--fold pair passes. Thus, GeoLASR summarizes typical multiplicative relative risk, while WorstLASR exposes the least favorable pair.

\subsection{Theoretical Interpretation}
\paragraph{Conditional support beyond the state.} Assume $\mathbb E[\|Y\|_2^2]<\infty$, and let $m_S=\mathbb E[Y\mid S]$ and $m_{S,Z}=\mathbb E[Y\mid S,Z]$. We consider prediction under squared loss, for which the Bayes predictor is the conditional mean~\citep{banerjee2005clustering}. For an input variable $V$, we define the optimal Bayes prediction risk under squared loss as
\begin{equation}
 R^*(V)=\inf_f\mathbb E[\|Y-f(V)\|_2^2].
 \label{eq:bayes-risk}
\end{equation}
Then, the reduction in optimal risk obtained by augmenting $S$ with an additional $Z$ admits the standard variance decomposition:
\begin{equation}
 R^*(S)-R^*(S,Z)
 =\mathbb E[\|m_{S,Z}-m_S\|_2^2].
 \label{eq:risk-decomposition}
\end{equation}
Equation~\eqref{eq:risk-decomposition} provides a predictive interpretation of the additional information carried by $Z$ beyond $S$: the risk reduction equals the expected squared distance between the conditional means with and without conditioning on $Z$. Formal statements and proofs are provided in Appendix Sec.~B.2.

\paragraph{When action information is needed.} Applying Equation~\eqref{eq:risk-decomposition} to the recorded action $A_t$ and an action-related target $Y_\tau$ gives
\begin{equation}
 R^*(S_t)-R^*(S_t,A_t)
 =\mathbb E\!\left[
 \left\|\mathbb E[Y_\tau\mid S_t,A_t]
 -\mathbb E[Y_\tau\mid S_t]\right\|_2^2
 \right].
 \label{eq:action-necessity}
\end{equation}
The right-hand side is positive if and only if the two conditional means differ with positive probability. This identity motivates a direct empirical check: for the chosen target, the recorded-action reference should reduce held-out prediction risk relative to the state route.
LASR imposes a stricter calibration condition. We interpret learned-representation scores only after confirming that $\operatorname{WorstLASR}(A;\Omega)<1$, which means that the recorded-action route outperforms both reference routes for every target--fold pair in $\Omega$.

\paragraph{Retention of action-relevant predictive information.}
Consider an idealized setting in which $B=(S,N)$ denotes the full context available at decision time, where $S$ is the state and $N$ denotes non-action variables. Let $A$ denote the recorded action, and let $Z=q(B,A)$ be a deterministic representation of the action and context. For any loss $L$ with finite Bayes risks, let $R_L^*(V)=\inf_f\mathbb E[L(Y,f(V))]$. The fact that $Z=q(B,A)$ means that any predictor using $(B,Z)$ can be reproduced from $(B,A)$, while a predictor using only $B$ is a special case of one using $(B,Z)$. Bayes-risk monotonicity therefore gives the first relation below; the second normalizes the risk reduction provided by $Z$ by that provided by $A$ to define $Q_{\mathrm{oracle}}$:
\begin{align}
 R_L^*(B,A)&\leq R_L^*(B,Z)\leq R_L^*(B),
 \label{eq:oracle-sandwich}\\
 Q_{\mathrm{oracle}}(Z)
 &=\frac{R_L^*(B)-R_L^*(B,Z)}
 {R_L^*(B)-R_L^*(B,A)}\in[0,1],
 \label{eq:oracle-retention}
\end{align}
where the second line requires $R_L^*(B)>R_L^*(B,A)$. The first line places the latent risk between access to the recorded action and access to non-action context alone; the second normalizes the predictive value retained by $Z$ relative to the recorded action. Under logarithmic loss, when the relevant conditional mutual informations are finite, $Q_{\mathrm{oracle}}(Z)=I(Y;Z\mid B)/I(Y;A\mid B)$, and under the same conditions $Q_{\mathrm{oracle}}(Z)=1$ if and only if $I(Y;A\mid B,Z)=0$. Thus, $Q_{\mathrm{oracle}}$ describes retained action-relevant predictive information in this idealized setting, not executability. The action-necessity, oracle-sufficiency, and information-retention proofs are provided in Appendix Sec.~B.3.
\subsection{Instantiating LASR as a Benchmark}

The LASR metric and the preceding analysis are independent of any particular dataset. Applying it to a frozen representation nevertheless requires a synchronized audit set with the state and action-related labels needed to define the three routes and targets. We next describe the common choices used to instantiate LASR as a benchmark on ALOHA. Every representation is evaluated using the same aligned transitions, targets, episode splits, affine ridge predictor family, mean-squared loss, candidate controls, and target--fold pairs. 

Each latent $Z_t$ is aligned with $(S_t,A_t,S_{t+1})$. We use the two action-related targets defined in the Evaluation Task. For each target and fold, the latent route is compared with the state route and a control selected from episode progress and episode identity on validation data. Five folds and two targets define the ten pairs in $\Omega$. The recorded action serves as a positive calibration reference. The same benchmark is used to compare frozen representations and to evaluate the adapted representation introduced next.

\subsection{LASR-Guided Residual Adapter}

\paragraph{Post-hoc accessibility cannot create information.} Let a fixed deterministic adapter produce $H=g(Z)$ from $Z$ alone at test time. Then
\begin{equation}
I(Y;H\mid B)\leq I(Y;Z\mid B),
\qquad
R_L^*(B,H)\geq R_L^*(B,Z).
\label{eq:adapter-boundary}
\end{equation}
The two inequalities state that the adapter cannot add target information or achieve lower Bayes risk than direct access to $Z$. It can nevertheless make existing information easier to access for a restricted predictor. A lower empirical LASR score after adaptation should therefore be interpreted as improved accessibility under the evaluation predictor, not as creation of new action information. This boundary clarifies the interpretation of the LASR-guided residual adapter described below.

LASR remains a held-out evaluation metric. We additionally use a LASR-inspired surrogate to train a small adapter that makes action-related information in a frozen latent easier for the evaluation predictor to access. Given a raw latent $Z_{\mathrm{raw}}$, the residual adapter produces
\begin{equation}
 H=Z_{\mathrm{raw}}+\Delta_{\theta}(Z_{\mathrm{raw}}).
 \label{eq:adapter-map}
\end{equation}
At test time, the adapter receives only $Z_{\mathrm{raw}}$. The post-hoc information boundary is proved in Appendix Sec.~B.4, and architecture and capacity settings are given in Appendix Sec.~F.

To focus on target variation not explained by the state, a state predictor fitted and frozen on training episodes defines
\begin{equation}
 r_j=Y_j-\widehat{\mathbb E}[Y_j\mid S].
 \label{eq:adapter-residual-target}
\end{equation}
An affine head predicts the standardized residual from $H$. Let $\mathcal L^{\mathrm{train}}_j(H)$ denote its mean-squared training loss. Using fixed state and progress references, we define
\begin{align}
 D_j
 &=\min\{\mathcal L^{\mathrm{train}}_{\mathrm{state},j},
 \mathcal L^{\mathrm{train}}_{\mathrm{progress},j}\}, \nonumber\\
 \widetilde\rho_j(H)
 &=\frac{\mathcal L^{\mathrm{train}}_j(H)+\varepsilon_{\mathrm{opt}}}
 {D_j+\varepsilon_{\mathrm{opt}}}.
 \label{eq:adapter-surrogate-ratio}
\end{align}
This ratio is used only to train the adapter; the resulting representation is evaluated with the held-out LASR metric defined above.

For $K$ training targets, the adapter minimizes
\begin{align}
 \mathcal L_{\mathrm{adapter}}
 &=\frac{1}{K}\sum_{j=1}^{K}\log\widetilde\rho_j(H)
 \nonumber\\
 &\quad+\beta\gamma\log\sum_{j=1}^{K}
 \exp\!\left(\frac{\log\widetilde\rho_j(H)}{\gamma}\right)
 \nonumber\\
 &\quad+\lambda_{\mathrm{id}}\|H-Z_{\mathrm{raw}}\|_2^2.
 \label{eq:adapter-objective}
\end{align}
The first term lowers the average log ratio, the second emphasizes the largest ratio, and the third penalizes large changes to the raw latent. The constant $\varepsilon_{\mathrm{opt}}$ is used only during training.

The adapter is trained on training episodes, selected on validation episodes, and evaluated from scratch on held-out test episodes using the same LASR procedure as all other representations.

We evaluate three adapter modes. The residual mode adds a learned correction to $Z_{\mathrm{raw}}$; replacement removes the raw-latent skip connection and uses the adapter output alone; augmentation concatenates the raw and adapter-produced representations. Their architectural, training, and uncertainty details are provided in Appendix Sec.~F. A lower held-out LASR after adaptation indicates that action-related information is more accessible to the evaluation predictor.

\begin{table*}[!t]
\centering
\small
\renewcommand{\arraystretch}{1.05}
\setlength{\tabcolsep}{5pt}
\begin{tabular}{@{}lrrrrrr@{}}
\hline
& & \multicolumn{2}{c}{\textbf{Original ALOHA panel}} &
\multicolumn{3}{c}{\textbf{Matched multi-task panel}} \\
\textbf{Representation} & \textbf{ALOHA dim.} &
\textbf{Geo$\downarrow$} & \textbf{Worst$\downarrow$} &
\textbf{Geo$\downarrow$} & \textbf{Worst$\downarrow$} &
\textbf{Branch acc. (\%)$\uparrow$} \\
\hline
DiLA $z$ & 256 & \textbf{0.750} & \textbf{0.796} & \textbf{0.688} & \textbf{0.719} & \textbf{65.58} \\
LaWAM $\mu_z$ & 32 & 0.891 & 0.947 & 0.860 & 0.870 & 48.59 \\
LAPA LAQ & 32 & 0.922 & 0.966 & 0.831 & 0.849 & 50.39 \\
UniVLA $z_q$ & 512 & 1.009 & 1.077 & 1.010 & 1.049 & 25.44 \\
MVP-LAM $z_q$ & 512 & 1.095 & 1.157 & 0.988 & 0.993 & 26.75 \\
\hline
Recorded action$^\dagger$ & 14 & 0 & 0.174 & $\approx0$ & 0.002 & N/A \\
\hline
\end{tabular}
\caption{
\textbf{LASR calibration and model comparison.}
Original ALOHA and matched-panel results; full uncertainty and target--fold
results are in Appendix Sec.~H.1. Lower LASR is better; one is the
operational boundary. Matched learned routes use eight-dimensional train-only
projections. Bold marks the best ranked learned point estimates.
$^\dagger$Unranked positive reference. Branch accuracy is common-decoder
one-step physical replay (state-only/chance: 25\%) and is not reported for the
recorded-action reference. Further audits are in Appendix Secs.~H.1--H.6.
}
\label{tab:model-calibration}
\end{table*}

\section{Experiments}

We retain recorded-action calibration, five frozen model representations, controlled DiLA transformations, and LASR-guided adapter evaluations. Table~\ref{tab:model-calibration} combines ALOHA and matched multi-task results; Table~\ref{tab:dila-adapter-prediction} reports diagnostics, adapter ablations, and frozen world-model prediction.

\subsection{Benchmark Data and Evaluation Protocol}

We evaluate 49 synchronized ALOHA static-battery demonstrations~\citep{zhao2023learning}, using a fixed set of 2,048 aligned transitions shared by every representation (Appendix Sec.~C).

We additionally evaluate a matched five-task ManiSkill3 panel~\citep{tao2024maniskill3}. Each exact simulator state anchors four action branches, with all branches kept in the same fan-disjoint split. Targets are the recorded 7D action and measured controllable TCP effect (Appendix Sec.~H.1).

We compare five frozen models: DiLA~\citep{zhang2026dila}, UniVLA~\citep{bu2025univla}, MVP-LAM~\citep{lee2026mvp}, LAPA~\citep{ye2025latent}, and LaWAM~\citep{chen2026lawam}. Their representations are aligned to the same transitions and evaluated on the 14-dimensional action and state-change targets; extraction rules and matched preprocessing appear in Appendix Secs.~D and H.1.

We use five episode-disjoint folds: three for predictor fitting, one for validation, and one for held-out testing. All routes use the same affine ridge family; validation selects the control before one test evaluation. Transformations are train-only, while adapter settings and stopping are validation-selected before from-scratch held-out evaluation. Paired episode-level bootstrap intervals are used for adapter contrasts; other comparisons are descriptive point estimates. All representations share the same transitions and valid set \(\Omega\) (Appendix Secs.~C--F and H.1). Computing environments and code-artifact scope are documented in Appendix Sec.~I.

\subsection{LASR Calibration and Model Comparison}

Table~\ref{tab:model-calibration} uses the recorded action as a positive reference to test whether LASR detects known action information. Its WorstLASR is below one, confirming that the recorded-action route outperforms both reference routes for both targets in all five held-out folds.

The remaining rows compare five frozen representations. The ALOHA columns retain the native-contract comparison; the matched columns place learned routes under a common width and protocol. The separation between strongly supported and boundary-level routes remains, and branch decoding follows the same ordering. Pair-only, task-stratified, uncertainty, and target--fold results are in Appendix Sec.~H.1.

The five-model GeoLASR ranking matches the held-out one-step branch-accuracy ranking. Across 25 model--task cells, lower GeoLASR is associated with lower executed TCP-effect MSE (task-block Spearman $\rho=0.956$) and lower branch error ($\rho=0.810$; Appendix Sec.~H.4). On official UniVLA rollouts, the native hidden route passes all LASR cells for a high-success policy, whereas random and shuffled controls fail. Together, these results support LASR as a representation-level admission metric with downstream behavioral relevance under controlled one-step execution; Appendix Sec.~H.5 provides a complementary distribution-matched native-policy audit.

Positive state-only gain can be rejected by a stronger non-action control (Appendix Sec.~H.3). Matched probes and expanded controls preserve the broad conclusion (Appendix Sec.~H.2), while camera and outcome interventions characterize robustness and sensitivity under controlled shifts (Appendix Sec.~H.6).

\begin{table}[t]
\centering
\small
\renewcommand{\arraystretch}{0.90}
\setlength{\tabcolsep}{2.0pt}
\begin{tabular}{@{}lrrr@{}}
\multicolumn{4}{c}{\textbf{(a) Representation diagnostics}} \\
\hline
\textbf{Variant} &
\textbf{Dim.} &
\textbf{GeoLASR$\downarrow$} &
\textbf{WorstLASR$\downarrow$} \\
\hline
Native DiLA & 256 & \textbf{0.750} & \textbf{0.796} \\
PCA & 16 & 0.954 & 0.980 \\
Bottleneck & 16 & 0.849 & 0.877 \\
Bottleneck & 8 & 0.913 & 0.949 \\
Progress control & 1 & 1.000 & 1.000 \\
Recorded action$^\dagger$ & 14 & 0 & 0.174 \\
\hline
\end{tabular}

\par\smallskip

\begin{tabular}{@{}lrr@{}}
\multicolumn{3}{c}{\textbf{(b) Adapter ablation}} \\
\hline
\textbf{Adapter} &
\textbf{GeoLASR$\downarrow$} &
\textbf{WorstLASR$\downarrow$} \\
\hline
Identity & 0.750 & 0.796 \\
LASR residual & \textbf{0.745} & \textbf{0.785} \\
MSE residual & 0.759 & 0.810 \\
Random residual & 0.762 & 0.815 \\
Shuffled residual & 0.761 & 0.834 \\
LASR replacement & 0.770 & 0.855 \\
LASR augmentation & 0.749 & 0.834 \\
\hline
\end{tabular}

\par\smallskip

\begin{tabular}{@{}llrrr@{}}
\multicolumn{5}{c}{\textbf{(c) World-model prediction}} \\
\hline
\textbf{Model} &
\textbf{Step} &
\textbf{DINO$\downarrow$} &
\textbf{SSIM$\uparrow$} &
\textbf{LPIPS$\downarrow$} \\
\hline
Original DiLA & 1 & 0.389 & 0.818 & 0.302 \\
\textbf{Adapted DiLA} & 1 &
\textbf{0.378} & \textbf{0.824} & \textbf{0.288} \\
Original DiLA & 15 & 0.521 & 0.828 & 0.339 \\
\textbf{Adapted DiLA} & 15 &
\textbf{0.494} & \textbf{0.830} & \textbf{0.324} \\
\hline
\end{tabular}
\caption{
\textbf{DiLA diagnostics, adapter ablations, and world-model prediction.}
Identity denotes unmodified DiLA; the model in (c) uses the LASR residual adapter in (b).
Arrows show preferred directions, and bold marks the best non-oracle point estimates.
$^\dagger$ Recorded-action GeoLASR is $1.19{\times}10^{-7}$ (shown as 0).
}
\label{tab:dila-adapter-prediction}
\end{table}

\subsection{Representation Diagnostics}

Table~\ref{tab:dila-adapter-prediction}(a) compares native DiLA with transformations of the same representation. PCA moves GeoLASR and WorstLASR toward one. At matched output size, the learned bottleneck retains stronger support than PCA, while its smaller version is weaker. Thus, LASR depends on preserved accessible action structure, not dimension alone; full sensitivity results are in Appendix Sec.~E.

The frozen-stratum analysis preserves the main ordering. Ridge and dimension checks preserve the below-one status of the strongest representations, with larger changes near the boundary and at high dimension.

\subsection{LASR-Guided Adapter and World-Model Prediction}

Table~\ref{tab:dila-adapter-prediction}(b) compares the LASR-guided residual adapter with the original representation and several adapter alternatives. All variants are evaluated from scratch using the same held-out LASR procedure. The LASR-guided residual adapter has the lowest GeoLASR and WorstLASR point estimates. Paired comparisons favor it over the MSE-, random-, and shuffled-supervision residual adapters on both summaries. Because these variants use the same residual form but different training signals, the comparison supports the use of the LASR-inspired surrogate. The residual mode also gives the best point estimates among the tested LASR adapter modes. Relative to the original representation, it improves both LASR summaries at the point-estimate level, with the paired interval spanning zero. Architecture, training, complete ablations, and paired uncertainty are reported in Appendix Sec.~F.

Table~\ref{tab:dila-adapter-prediction}(c) evaluates the same LASR-guided residual adapter in frozen world-model prediction. The adapted representation has better DINO, SSIM, and LPIPS point estimates at both evaluated horizons. The consistent direction across metrics and horizons provides descriptive evidence that the adapted representation also supports better world-model prediction beyond the LASR evaluation. Per-horizon paired results and implementation details are reported in Appendix Sec.~G.

\section{Discussion and Limitations}

LASR provides a controlled offline metric specifically for action-grounded predictive support. In the fixed ALOHA setting, its dimensionless scores support graded comparisons, while WorstLASR indicates whether support is consistent across all evaluated pairs. The compression and adapter results are consistent with a central interpretation: LASR reflects how accessible action-related predictive structure is under the chosen predictor family. The accompanying world-model prediction point estimates suggest that this accessibility may be relevant beyond the evaluation predictor, while paired adapter comparisons characterize uncertainty.

LASR remains controlled and interpretable by fixing the targets, predictor family, and controls. The matched multi-task and nonlinear-probe audits (Appendix Secs.~H.1--H.2), physical replay (Appendix Sec.~H.4), and registered interventions (Appendix Sec.~H.6) broaden the original ALOHA evidence without changing the metric definition. The result remains relative to the declared audit labels, probe family, and controls, and low LASR alone does not certify causal action identity or executability.

\section{Conclusion}

We introduced LASR, a dimensionless held-out metric that compares matched state, latent, and validation-selected control routes to measure whether a learned latent supports action-related targets beyond state and non-action alternatives. GeoLASR summarizes typical support, while WorstLASR exposes the least favorable target--fold result. On ALOHA, LASR detects the predictive support of recorded action, requires learned representations to outperform validation-selected non-action explanations, and provides graded comparisons of learned and transformed representations. Matched multi-task, probe-family, physical-replay, and intervention audits broaden this interpretation across declared evaluation conditions. Beyond evaluation, the LASR-guided residual adapter attains the best LASR point estimates among the tested adapter variants and consistently improves DINO, SSIM, and LPIPS point estimates at the evaluated prediction horizons. Paired comparisons further favor LASR supervision over MSE, random, and shuffled supervision. LASR therefore connects prediction improvement to action-grounded latent evaluation through a clearly scoped offline protocol.

\clearpage
\appendix
\section{Definitions and Main Statements}

The first two sections define LASR and give the proofs referenced in the main paper. The remaining sections describe the ALOHA transition panel, model-specific representation extraction, transformation checks, adapter training, uncertainty estimates, world-model prediction, and the extended matched-panel, physical-replay, native-policy, and intervention audits.

For each target $\tau$ and held-out fold $k$, LASR compares three predictors on the same held-out data: one uses the state, one adds the latent, and one adds a non-action control. This comparison produces one per-target--fold score for each representation. The state comparison follows the conditional-probing principle of asking what becomes usable beyond a baseline~\citep{hewitt2021conditional}. The non-action route serves a distinct purpose from standard probe controls~\citep{hewitt2019designing}: it tests a concrete trajectory alternative under the same prediction task.

\paragraph{Per-target--fold score.} Fix a target $\tau$ and held-out fold $k$, with current state $S$, latent representation $Z$, target $Y_\tau$, and fixed family $\mathcal C$ of candidate non-action controls. Select
\begin{equation}
 b^*_{\tau,k}
 =\arg\min_{b\in\mathcal C}
 \widehat R^{\mathrm{val}}_{\tau,k}(S,b),
 \label{eq:lasr-control-selection}
\end{equation}
then evaluate the selected control once on test. The separately fitted state, latent, and control predictors yield three empirical held-out prediction risks, each defined as the common prediction loss averaged over the test transitions in fold $k$. In the ALOHA benchmark, the common loss is mean-squared loss. Let
\begin{align}
 E_{\mathrm{state},\tau,k} &= \widehat R^{\mathrm{test}}_{\tau,k}(S),\\
 E_{Z,\tau,k} &= \widehat R^{\mathrm{test}}_{\tau,k}(S,Z),\\
 E_{\mathrm{ctrl},\tau,k} &=
 \widehat R^{\mathrm{test}}_{\tau,k}(S,b^*_{\tau,k}).
\end{align}
We suppress the indices $(\tau,k)$ when the comparison is clear. For finite, nonnegative $E_Z$ and finite, strictly positive $E_{\mathrm{state}}$ and $E_{\mathrm{ctrl}}$, define
\begin{equation}
 \rho_{\tau,k}(Z)
 =\frac{E_Z}{\min\{E_{\mathrm{state}},E_{\mathrm{ctrl}}\}}.
 \label{eq:lasr-target}
\end{equation}
The numerator is the latent-route error. The denominator is the smaller, and therefore stronger, of the two reference-route errors. A score below one means that the latent route beats both references. No clipping or smoothing is applied. If either common reference-route risk is nonpositive or nonfinite, the score for that index pair is undefined and is handled by the common-index-set rule below.

\begin{proposition}[Threshold interpretation and scale invariance]
\label{prop:lasr-separation}
For every valid target--fold pair $(\tau,k)$, we have
\begin{equation}
 \rho_{\tau,k}(Z)<1
 \quad\Longleftrightarrow\quad
 E_Z<E_{\mathrm{state}}
 \ \text{and}\ 
 E_Z<E_{\mathrm{ctrl}}.
 \label{eq:lasr-conjunction}
\end{equation}
Moreover, we have
\begin{equation}
 1-\rho_{\tau,k}(Z)
 =\min\left\{
 1-\frac{E_Z}{E_{\mathrm{state}}},
 1-\frac{E_Z}{E_{\mathrm{ctrl}}}
 \right\}.
 \label{eq:lasr-conservative-improvement}
\end{equation}
A common positive rescaling of all three risks leaves LASR unchanged.
\end{proposition}

\paragraph{Common valid pairs and aggregate scores.} The planned index set is $\mathcal T\times\mathcal F$. Let $\Omega\subseteq\mathcal T\times\mathcal F$ denote the fixed common set of target--fold pairs for which the shared state and control risks are finite and strictly positive. The same $\Omega$ is used for every representation in one ranking. If a shared reference risk is invalid, that pair is excluded for every representation and the exclusion is disclosed. If $E_Z$ is nonfinite or negative for only one representation on the fixed $\Omega$, that representation is marked invalid and reported separately; $\Omega$ is not reduced for that representation. In the ALOHA benchmark, all two-target, five-fold pairs are valid, so $\Omega=\mathcal T\times\mathcal F$ and $|\Omega|=10$.

For a representation that is valid on all of $\Omega$, define
\begin{align}
 \operatorname{GeoLASR}(Z;\Omega)
 &=\left(
 \prod_{(\tau,k)\in\Omega}\rho_{\tau,k}(Z)
 \right)^{1/|\Omega|},
 \label{eq:lasr-geo}\\
 \operatorname{WorstLASR}(Z;\Omega)
 &=\max_{(\tau,k)\in\Omega}\rho_{\tau,k}(Z).
 \label{eq:lasr-worst}
\end{align}
Both summaries are lower-is-better, and one is the operational boundary. GeoLASR summarizes typical multiplicative support; for strictly positive scores, its logarithm is the mean log ratio. WorstLASR is the largest score and exposes the least favorable target--fold result.

\paragraph{Population quantities used in the theory.} For a feature set $V$, the population Bayes squared risk is
\begin{equation}
 R^*(V)=\inf_f \mathbb E[\|Y-f(V)\|_2^2].
 \label{eq:bayes-squared-risk}
\end{equation}

\begin{proposition}[Conditional-risk decomposition]
\label{prop:lasr-risk-decomposition}
Let $m_S=\mathbb E[Y\mid S]$ and $m_{S,Z}=\mathbb E[Y\mid S,Z]$. Then we have
\begin{equation}
 R^*(S)-R^*(S,Z)
 =\mathbb E[\|m_{S,Z}-m_S\|_2^2].
 \label{eq:lasr-risk-decomposition}
\end{equation}
When $R^*(S)>0$, we have
\begin{equation}
 \frac{R^*(S,Z)}{R^*(S)}
 =1-\frac{\mathbb E[\|m_{S,Z}-m_S\|_2^2]}{R^*(S)}.
 \label{eq:lasr-state-explained-fraction}
\end{equation}
\end{proposition}

For $B=(S,C)$, an additional nested diagnostic is
\begin{equation}
 \rho_{\mathrm{cond}}(Z;C)=\frac{R^*(S,C,Z)}{R^*(S,C)},
 \label{eq:lasr-nested}
\end{equation}
and Bayes-risk monotonicity gives
\begin{equation}
 \rho_{\mathrm{cond}}(Z;C)
 \leq \frac{R^*(S,Z)}{R^*(S,C)}.
 \label{eq:lasr-nested-bound}
\end{equation}

\begin{proposition}[When action information is necessary]
\label{prop:lasr-action-necessity}
For recorded action $A_t$ and a square-integrable target $Y_\tau$, we have
\begin{equation}
\begin{aligned}
 R^*(S_t)-R^*(S_t,A_t)
 &=\mathbb E\Big[
 \big\|\mathbb E[Y_\tau\mid S_t,A_t] \\
 &\qquad-\mathbb E[Y_\tau\mid S_t]\big\|_2^2
 \Big].
\end{aligned}
 \label{eq:lasr-action-necessity}
\end{equation}
\end{proposition}

\begin{theorem}[Oracle sandwich and target-relevant action sufficiency]
\label{thm:lasr-oracle-sandwich}
Assume $Z=q(B,A)$ is a deterministic action abstraction. For any loss $L$ with well-defined Bayes risks, let $R_L^*(V)=\inf_f\mathbb E[L(Y,f(V))]$. Then we have
\begin{equation}
 R_L^*(B,A)\leq R_L^*(B,Z)\leq R_L^*(B).
 \label{eq:lasr-oracle-sandwich}
\end{equation}
If $R_L^*(B)>R_L^*(B,A)$, define
\begin{equation}
 Q_{\mathrm{oracle}}(Z)
 =\frac{R_L^*(B)-R_L^*(B,Z)}{R_L^*(B)-R_L^*(B,A)}.
 \label{eq:lasr-oracle-retention}
\end{equation}
Then $Q_{\mathrm{oracle}}(Z)\in[0,1]$. Under logarithmic loss and a rich probabilistic predictor, we have
\begin{equation}
 Q_{\mathrm{oracle}}(Z)
 =\frac{I(Y;Z\mid B)}{I(Y;A\mid B)}.
 \label{eq:lasr-oracle-information}
\end{equation}
Moreover, $Q_{\mathrm{oracle}}(Z)=1$ if and only if $Y\perp A\mid(B,Z)$, up to null sets.
\end{theorem}

\begin{corollary}[Post-hoc adapter boundary]
\label{cor:lasr-adapter-boundary}
Let $H=g(Z)$ be a deterministic post-hoc transformation of $Z$ that receives no additional test-time input. For any context $B$ and any loss $L$ with well-defined Bayes risks, we have
\begin{equation}
 I(Y;H\mid B)\leq I(Y;Z\mid B),
 \qquad R_L^*(B,H)\geq R_L^*(B,Z).
 \label{eq:lasr-adapter-boundary}
\end{equation}
\end{corollary}

\section{Proofs}

\subsection{Algebraic Properties of the LASR Score}

\noindent\textbf{Main-paper location. Equation~(1)} in the \emph{LASR Score} subsection defines
\begin{equation*}
 \rho_{\tau,k}(Z)=\frac{E_Z}{\min\{E_{\mathrm{state}},E_{\mathrm{ctrl}}\}}.
\end{equation*}
Proposition~\ref{prop:lasr-separation} formalizes the threshold interpretation and scale invariance stated immediately after that equation.

\begin{proof}[Proof of Proposition~\ref{prop:lasr-separation}]
Let $D=\min\{E_{\mathrm{state}},E_{\mathrm{ctrl}}\}>0$. Then $\rho_{\tau,k}(Z)=E_Z/D$. Hence $\rho_{\tau,k}(Z)<1$ if and only if $E_Z<D$, which is equivalent to $E_Z<E_{\mathrm{state}}$ and $E_Z<E_{\mathrm{ctrl}}$. For the improvement identity, use $1-\max\{a,b\}=\min\{1-a,1-b\}$ with $a=E_Z/E_{\mathrm{state}}$ and $b=E_Z/E_{\mathrm{ctrl}}$. Finally, if all risks are multiplied by $a_0>0$, then $(a_0E_Z)/\min\{a_0E_{\mathrm{state}},a_0E_{\mathrm{ctrl}}\}=E_Z/D$.
\end{proof}

\begin{proposition}[Aggregation bounds]
\label{prop:lasr-aggregation-bounds}
For positive per-target--fold LASR scores $\rho_1,\ldots,\rho_m$, let $G=(\prod_j\rho_j)^{1/m}$ and $W=\max_j\rho_j$. Then we have
\begin{equation}
 \min_j\rho_j\leq G\leq W,
 \qquad W<1\Longleftrightarrow \rho_j<1\ \text{for all }j.
 \label{eq:lasr-aggregation-bounds}
\end{equation}
Furthermore, we have
\begin{equation}
 \log G=\frac{1}{m}\sum_{j=1}^m\log\rho_j,
 \qquad
 \frac{\partial\log G}{\partial\log\rho_j}=\frac{1}{m}.
 \label{eq:lasr-log-sensitivity}
\end{equation}
\end{proposition}

\begin{proof}
Monotonicity of the logarithm gives $\min_j\log\rho_j\leq m^{-1}\sum_j\log\rho_j\leq \max_j\log\rho_j$. Exponentiating proves the first inequality. The equivalence for $W$ follows from the definition of a maximum. The two log identities follow by taking logarithms and differentiating. If any valid $\rho_j$ is zero, the product convention sets $G=0$; the WorstLASR certificate remains unchanged and prevents a zero on one target from accepting a representation that fails another target.
\end{proof}

\begin{proposition}[Population control-family monotonicity]
\label{prop:lasr-control-monotonicity}
Define the population control risk $E_{\mathrm{ctrl}}(\mathcal C)=\inf_{b\in\mathcal C}R(S,b)$. If $\mathcal C\subseteq\mathcal C'$, then, for fixed $E_Z$ and $E_{\mathrm{state}}$, we have
\begin{equation}
 \rho_{\mathcal C'}(Z)\geq\rho_{\mathcal C}(Z).
 \label{eq:lasr-control-monotonicity}
\end{equation}
\end{proposition}

\begin{proof}
Taking an infimum over a superset cannot increase the infimum, so $E_{\mathrm{ctrl}}(\mathcal C')\leq E_{\mathrm{ctrl}}(\mathcal C)$. The denominator $\min\{E_{\mathrm{state}},E_{\mathrm{ctrl}}\}$ therefore cannot increase, and the LASR score cannot decrease.
\end{proof}

Proposition~\ref{prop:lasr-control-monotonicity} is a population statement. In the finite protocol, Eq.~\eqref{eq:lasr-control-selection} selects a control on validation and evaluates it on test. Enlarging the candidate set can change the selected control and may increase or decrease its realized held-out risk. Deterministic finite-sample monotonicity must therefore not be claimed; the selected control identity and all validation and held-out risks are retained.

\subsection{Squared-Risk Decomposition}

\noindent\textbf{Main-paper location. Equation~(5)} in the \emph{Conditional support beyond the state} paragraph of the \emph{Theoretical Interpretation} subsection states
\begin{equation*}
 R^*(S)-R^*(S,Z)
 =\mathbb E\!\left[\|m_{S,Z}-m_S\|_2^2\right].
\end{equation*}
Proposition~\ref{prop:lasr-risk-decomposition} gives the formal version of this identity.

\begin{proof}[Proof of Proposition~\ref{prop:lasr-risk-decomposition}]
Let $m_S=\mathbb E[Y\mid S]$ and $m_{S,Z}=\mathbb E[Y\mid S,Z]$. Write
\begin{equation}
 Y-m_S=(Y-m_{S,Z})+(m_{S,Z}-m_S).
\end{equation}
After expanding the squared norm, the cross term has expectation zero:
\begin{align}
 &\mathbb E[(Y-m_{S,Z})^{\top}(m_{S,Z}-m_S)]\\
 &\quad=\mathbb E[\mathbb E[Y-m_{S,Z}\mid S,Z]^{\top}
 (m_{S,Z}-m_S)]=0.
\end{align}
The conditional mean minimizes expected squared risk, so $R^*(S)=\mathbb E\|Y-m_S\|_2^2$ and $R^*(S,Z)=\mathbb E\|Y-m_{S,Z}\|_2^2$. Substitution yields Eq.~\eqref{eq:lasr-risk-decomposition}; division by $R^*(S)>0$ gives Eq.~\eqref{eq:lasr-state-explained-fraction}.
\end{proof}

The same argument with $B=(S,C)$ yields
\begin{equation}
 R^*(B)-R^*(B,Z)
 =\mathbb E\left[
 \left\|\mathbb E[Y\mid B,Z]-\mathbb E[Y\mid B]\right\|_2^2
 \right].
 \label{eq:lasr-nested-risk-decomposition}
\end{equation}
Since adding $C$ cannot increase Bayes risk, $R^*(S,C,Z)\leq R^*(S,Z)$, which gives Eq.~\eqref{eq:lasr-nested-bound}. The canonical nonnested ratio is therefore a control-route dominance requirement rather than an exact conditional-information estimate.

\subsection{Action Necessity and Oracle Sufficiency}

\noindent\textbf{Main-paper location. Equation~(6)} in the \emph{When action information is needed} paragraph of the \emph{Theoretical Interpretation} subsection states
\begin{equation*}
 R^*(S_t)-R^*(S_t,A_t)
 =\mathbb E\!\left[\left\|\mathbb E[Y_\tau\mid S_t,A_t]-\mathbb E[Y_\tau\mid S_t]\right\|_2^2\right].
\end{equation*}
Proposition~\ref{prop:lasr-action-necessity} establishes when this quantity is strictly positive.

\begin{proof}[Proof of Proposition~\ref{prop:lasr-action-necessity}]
Apply Proposition~\ref{prop:lasr-risk-decomposition} with $Z=A_t$. The right-hand side of Eq.~\eqref{eq:lasr-action-necessity} is nonnegative and is zero if and only if the two conditional means agree almost surely. It is strictly positive exactly when they differ on a set of nonzero probability.
\end{proof}

\noindent\textbf{Main-paper location. Equations~(7)--(8)} in the \emph{Retention of action-relevant predictive information} paragraph of the \emph{Theoretical Interpretation} subsection state
\begin{align*}
 R_L^*(B,A)&\leq R_L^*(B,Z)\leq R_L^*(B),\\
 Q_{\mathrm{oracle}}(Z)&=\frac{R_L^*(B)-R_L^*(B,Z)}{R_L^*(B)-R_L^*(B,A)}\in[0,1].
\end{align*}
Theorem~\ref{thm:lasr-oracle-sandwich} proves this ordering and the normalized retention result.

\begin{proof}[Proof of Theorem~\ref{thm:lasr-oracle-sandwich}]
Because $Z=q(B,A)$ is deterministic, $(B,Z)$ is a measurable function of $(B,A)$. Any decision rule based on $(B,Z)$ can be reproduced from $(B,A)$, while a rule based only on $B$ is available from $(B,Z)$ by ignoring $Z$. Taking the infimum of risk over these rule classes gives $R_L^*(B,A)\leq R_L^*(B,Z)\leq R_L^*(B)$, consistent with Blackwell's ordering of experiments \citep{blackwell1953equivalent}. Subtracting each term from $R_L^*(B)$ and dividing by the positive oracle improvement proves $Q_{\mathrm{oracle}}(Z)\in[0,1]$.

Under logarithmic loss, Bayes risk is conditional entropy: $R_L^*(B)=H(Y\mid B)$, $R_L^*(B,Z)=H(Y\mid B,Z)$, and $R_L^*(B,A)=H(Y\mid B,A)$. Therefore
\begin{align}
 R_L^*(B)-R_L^*(B,Z)&=I(Y;Z\mid B),\\
 R_L^*(B)-R_L^*(B,A)&=I(Y;A\mid B).
\end{align}
Since $Z$ is a function of $(B,A)$, the conditional chain rule gives
\begin{equation}
 I(Y;A\mid B)=I(Y;Z\mid B)+I(Y;A\mid B,Z).
\end{equation}
Equation~\eqref{eq:lasr-oracle-information} follows. The ratio equals one if and only if $I(Y;A\mid B,Z)=0$, equivalent to $Y\perp A\mid(B,Z)$ up to null sets. Strict propriety makes this a statement about the full conditional predictive distribution \citep{gneiting2007strictly}.
\end{proof}

\begin{corollary}[Linear-Gaussian action channel]
\label{cor:lasr-linear-gaussian}
Suppose, after conditioning on $B$, that
\begin{equation}
 Y=\mu(B)+\gamma A+\varepsilon,
 \qquad Z=A+\eta,
 \label{eq:lasr-linear-gaussian-scm}
\end{equation}
where $A\sim\mathcal N(0,\sigma_A^2)$, $\eta\sim\mathcal N(0,\sigma_\eta^2)$, and $\varepsilon\sim\mathcal N(0,\sigma_\varepsilon^2)$ are mutually independent. Under Bayes squared risk,
\begin{align}
 R^*(B)&=\gamma^2\sigma_A^2+\sigma_\varepsilon^2,\\
 R^*(B,Z)&=\sigma_\varepsilon^2+
 \gamma^2\frac{\sigma_A^2\sigma_\eta^2}
 {\sigma_A^2+\sigma_\eta^2},\\
 R^*(B,A)&=\sigma_\varepsilon^2,
 \label{eq:lasr-linear-gaussian-risks}
\end{align}
and
\begin{equation}
 Q_{\mathrm{oracle}}(Z)
 =\frac{\sigma_A^2}{\sigma_A^2+\sigma_\eta^2}.
 \label{eq:lasr-linear-gaussian-retention}
\end{equation}
The latent risk ratio increases strictly with $\sigma_\eta^2$ whenever $\gamma\neq0$ and converges to one as $\sigma_\eta^2\rightarrow\infty$.
\end{corollary}

\begin{proof}
The Gaussian conditional variance is
\begin{equation}
 \operatorname{Var}(A\mid Z)
 =\frac{\sigma_A^2\sigma_\eta^2}{\sigma_A^2+\sigma_\eta^2}.
\end{equation}
The three risks follow by conditioning on $B$, $(B,Z)$, and $(B,A)$. Subtracting the second and third risks from the first gives
\begin{align}
 R^*(B)-R^*(B,Z)
 &=\gamma^2\frac{\sigma_A^4}{\sigma_A^2+\sigma_\eta^2},\\
 R^*(B)-R^*(B,A)&=\gamma^2\sigma_A^2.
\end{align}
Their ratio is Eq.~\eqref{eq:lasr-linear-gaussian-retention}. Finally, $x\mapsto\sigma_A^2x/(\sigma_A^2+x)$ is strictly increasing for $x>0$. If $\gamma=0$, action has no conditional mean effect and all action-derived improvement terms are zero.
\end{proof}

\subsection{Post-hoc Accessibility}

\noindent\textbf{Main-paper location. Equation~(9)} in the \emph{Post-hoc accessibility cannot create information} paragraph of the \emph{LASR-Guided Residual Adapter} subsection states
\begin{equation*}
 I(Y;H\mid B)\leq I(Y;Z\mid B),
 \qquad R_L^*(B,H)\geq R_L^*(B,Z).
\end{equation*}
Corollary~\ref{cor:lasr-adapter-boundary} proves these two inequalities.

\begin{proof}[Proof of Corollary~\ref{cor:lasr-adapter-boundary}]
Because $H=g(Z)$ is deterministic, $(B,H)$ is a measurable function of $(B,Z)$ and $Y-(B,Z)-(B,H)$ forms a conditional Markov chain. The conditional data-processing inequality gives $I(Y;H\mid B)\leq I(Y;Z\mid B)$. Moreover, every decision rule based on $(B,H)$ can be reproduced from $(B,Z)$ by first computing $H=g(Z)$. Taking the infimum over the larger decision-rule class available from $(B,Z)$ therefore gives $R_L^*(B,H)\geq R_L^*(B,Z)$.
\end{proof}

\subsection{Finite-Sample Stability and Interpretation Boundaries}

\begin{proposition}[Ratio stability away from zero denominators]
\label{prop:lasr-ratio-stability}
Let $n=E_Z$, $d=\min\{E_{\mathrm{state}},E_{\mathrm{ctrl}}\}$, and $\rho=n/d$. Assume $d\geq\delta>0$, $n\leq M$, and every empirical risk is within $\xi<\delta$ of its population counterpart. Then we have
\begin{equation}
 |\widehat\rho-\rho|
 \leq \frac{\xi(\delta+M)}{\delta(\delta-\xi)}.
 \label{eq:lasr-ratio-stability}
\end{equation}
\end{proposition}

\begin{proof}
The minimum is one-Lipschitz under the maximum norm, so $|\widehat d-d|\leq\xi$ and $\widehat d\geq\delta-\xi$. Hence
\begin{align}
 \left|\frac{\widehat n}{\widehat d}-\frac{n}{d}\right|
 &\leq\frac{|\widehat n-n|}{\widehat d}
 +\frac{n|\widehat d-d|}{d\widehat d}\\
 &\leq\frac{\xi}{\delta-\xi}
 +\frac{M\xi}{\delta(\delta-\xi)},
\end{align}
which proves Eq.~\eqref{eq:lasr-ratio-stability}.
\end{proof}

This bound explains why the protocol invalidates nonpositive denominators and reports episode-cluster uncertainty. It does not establish finite-sample coverage after arbitrary model selection; fitting and selection remain frozen before test and bootstrap stages.

Two counterexamples delimit interpretation. If $Z=Y$ copies a future target, then $E_Z$ and LASR can be zero even though no action decoder or executable latent exists. If $Y\perp A\mid B$, even the recorded action has no conditional predictive advantage and its oracle LASR is one at the population optimum. Thus low LASR is not sufficient for executability, and a non-passing action oracle on a state-sufficient slice does not imply an implementation failure.

Finally, value-function guarantees for a learned world model require additional assumptions on reward prediction, transition-distribution error, coverage, Markov sufficiency, and an executable action interface \citep{gelada2019deepmdp}. LASR does not supply those assumptions and is not a substitute for a simulation lemma.

\section{ALOHA Data and LASR Evaluation Details}

This section gives the transition-selection, fold, control-selection, predictor, and validity details referenced by the main paper.

\paragraph{Dataset and eligible transitions.} We use all 49 episodes of the LeRobot v3.0 \texttt{lerobot/aloha\_static\_battery} release~\citep{zhao2023learning}. Each episode has 600 frames. A transition from $t$ to $t+1$ is eligible when $t\in\{14,\ldots,598\}$, both rows belong to the same episode, and the 14-dimensional state and action vectors are finite. The first 14 frames cannot provide the 16-frame context needed by DiLA, and the final frame has no successor. These rules leave $49\times585=28{,}665$ eligible transitions.

\begin{figure*}[t]
  \centering
  \includegraphics[width=\textwidth]{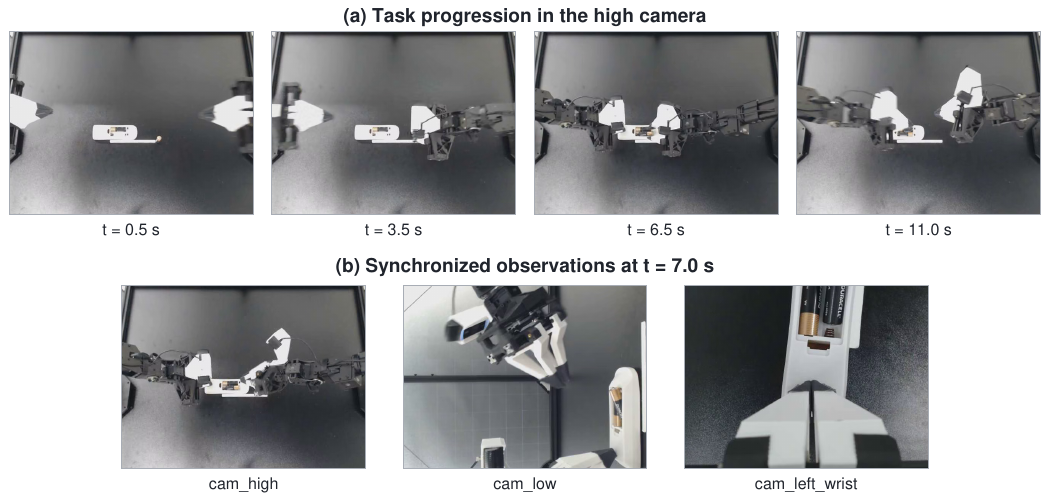}
  \caption{ALOHA data used to construct the analysis panel. (a) Four frames from a trajectory in the static-battery release illustrate task progression. (b) Synchronized observations from the high, low, and left-wrist cameras at one time step. All image-based representations evaluated in this paper are extracted from the aligned \texttt{cam\_high} stream.}
  \label{fig:supp-aloha-data}
\end{figure*}

We read $S_t$ from \texttt{observation.state}, $A_t$ from \texttt{action}, and $S_{t+1}$ from the next row. We do not interpolate, impute, or zero-fill any value. The metadata names the coordinates only as \texttt{motor\_0},\ldots,\texttt{motor\_13}; we therefore keep the recorded order and do not assign unverified joint names or units. Figure~\ref{fig:supp-aloha-data} illustrates the task progression and synchronized observations underlying these aligned transitions.

\paragraph{Panel composition.} The final panel contains 2,048 transitions. Half are \emph{ordinary-reference} transitions selected at approximately uniform temporal intervals within each episode under near-equal per-episode quotas (20 or 21 transitions per episode), retaining broad episode and temporal coverage. The other half are \emph{ranking-enriched} transitions that emphasize a harder case for the evaluator: two transitions from different episodes may be close in standardized state space under Euclidean distance but have different recorded actions. The two groups are interleaved, with 1,024 transitions from each group.

For the enriched half, we standardize the eligible state and action coordinates. A Euclidean state-space search over the full eligible transition pool considers $k=640$ neighbors and retains the 20 closest candidates from other episodes; every query contains at least 20 such candidates. We rank a candidate by
\begin{equation}
 q=\operatorname{pct}(d_A)\bigl(1-\operatorname{pct}(d_S)\bigr),
 \label{eq:supp-panel-rank}
\end{equation}
where $\operatorname{pct}(d)$ is the percentile rank of distance $d$. We keep candidates with $d_S\leq P_{50}$ and $d_A\geq P_{75}$, suppress nearby choices within 12 frames, and balance the number selected from each episode. We fix this selection rule before model evaluation. It uses recorded-action distance, but it does not use any learned latent, LASR score, or final model result.

\paragraph{Episode-disjoint folds.} For outer fold $f$, episodes with $e\bmod5=f$ form the test split. Episodes with $e\bmod5=(f+1)\bmod5$ form validation, and the remaining episodes form training. The five test splits contain 420, 418, 414, 420, and 376 transitions. Every episode is used as test data exactly once, and no episode appears in more than one split within a fold.

\paragraph{Non-action controls.} Progress is the scalar $c_{\mathrm{prog}}(t)=t/599$. Episode identity is a fixed 49-dimensional one-hot vector, with one coordinate for each panel episode. In a given fold, a held-out episode activates a coordinate that was always zero during training, so ridge regression assigns that coordinate a zero coefficient. Thus, the identity control cannot learn a coefficient specifically for an unseen test episode.

For each target--fold pair, we fit progress and episode identity on the training episodes, compare their validation MSE, and select the lower-risk control. The control name breaks an exact tie deterministically. Progress is selected in all ten target--fold comparisons, and the selected control is evaluated once on test.

\paragraph{Matched predictors and risk.} Each route has its own affine ridge predictor because the inputs differ, but all routes use the same predictor family, coefficient $10^{-6}$, loss, and fitting procedure. Feature means and standard deviations are estimated from training episodes only; a constant feature receives scale one. The intercept is not penalized. Targets remain in their original scale. We compute risk as MSE averaged over held-out transitions and all 14 target coordinates. The reported LASR score uses no clipping or denominator smoothing.

\section{Representation Extraction and Target-Specific Results}

All five representations come from public pretrained checkpoints and are frozen on ALOHA. The artifact records the public checkpoints and extraction settings. Every image-based route uses the same aligned \texttt{cam\_high} frames.

\paragraph{DiLA.} DiLA~\citep{zhang2026dila} receives the 16-frame context $t-14,\ldots,t+1$. Its inverse-dynamics module returns 15 latent vectors of width 256. We retain slot 14, which is aligned with the transition $t\rightarrow t+1$, as the continuous 256-dimensional representation \texttt{z\_raw}.

\paragraph{UniVLA.} UniVLA~\citep{bu2025univla} receives the frame pair $(t,t+1)$ scaled to $[0,1]$ and resized to $224\times224$. We apply \texttt{lam.vq\_encode}, retain the task-centric tensor of shape $1\times4\times128$, exclude the nuisance tensor, and flatten the result to 512 dimensions.

\paragraph{MVP-LAM.} MVP-LAM~\citep{lee2026mvp} uses the same frame pair resized to $224\times224$. We take the native $4\times128$ output of \texttt{vq\_encode} and flatten it to 512 dimensions.

\paragraph{LAPA.} LAPA~\citep{ye2025latent} returns four categorical slots, each with eight possible values. We one-hot encode each slot separately and concatenate the four vectors in their original order. The resulting 32-dimensional vector is a lossless encoding of the four tokens. The final panel contains 46 unique token sequences, and the most common sequence appears in $1429/2048\approx69.8\%$ transitions. We therefore mark LAPA as a low-diversity representation in the result tables.

\paragraph{LaWAM.} LaWAM~\citep{chen2026lawam} receives the frame pair resized to $256\times256$ with ImageNet normalization. We use its deterministic 32-dimensional VAE posterior mean and disable sampling. The released interface names this output \texttt{quantized}, but the returned value is not treated as a VQ code.

\begin{table}[!htbp]
\centering
\small
\setlength{\tabcolsep}{2.5pt}
\begin{tabular}{lcc}
\hline
& \textbf{Action target} & \textbf{State-change target} \\
\textbf{Representation} & \textbf{Geo (Worst)} & \textbf{Geo (Worst)} \\
\hline
DiLA & 0.735 (0.759) & 0.767 (0.796) \\
LaWAM & 0.887 (0.934) & 0.895 (0.947) \\
LAPA$^*$ & 0.921 (0.966) & 0.923 (0.956) \\
UniVLA & 1.011 (1.077) & 1.007 (1.066) \\
MVP-LAM & 1.098 (1.157) & 1.092 (1.137) \\
\hline
\end{tabular}
\caption{Target-specific summaries over five folds. Each entry is GeoLASR with WorstLASR in parentheses. The main paper combines the two targets into ten target--fold scores. $^*$LAPA retains the low-diversity warning described in the text.}
\label{tab:supp-target-results}
\end{table}

\paragraph{Results by panel half.} Table~\ref{tab:supp-strata} checks whether the ordering depends on the ordinary or enriched half of the panel. Predictor training and validation still use the complete fold. We change only the held-out rows used to compute each descriptive score.

\begin{table}[!htbp]
\centering
\small
\setlength{\tabcolsep}{2.5pt}
\begin{tabular}{lrrrr}
\hline
& \multicolumn{2}{c}{\textbf{Ordinary}} & \multicolumn{2}{c}{\textbf{Enriched}} \\
\textbf{Representation} & \textbf{Geo} & \textbf{Worst} & \textbf{Geo} & \textbf{Worst} \\
\hline
DiLA & 0.750 & 0.803 & 0.752 & 0.840 \\
LaWAM & 0.882 & 0.908 & 0.897 & 0.981 \\
LAPA$^*$ & 0.920 & 0.942 & 0.924 & 0.999 \\
UniVLA & 1.014 & 1.128 & 1.005 & 1.044 \\
MVP-LAM & 1.090 & 1.223 & 1.100 & 1.244 \\
\hline
\end{tabular}
\caption{Descriptive results for the two fixed panel halves. Predictors are still fitted on the full fold; only the held-out scoring rows are separated. These strata do not affect model selection or the main ranking.}
\label{tab:supp-strata}
\end{table}

\FloatBarrier
\section{Representation Transformations and Sensitivity Checks}

We test the sensitivity of LASR to representation compression, evaluator regularization, and representation dimensionality.

\paragraph{PCA.} The main PCA baseline reduces DiLA from 256 to 16 dimensions. For each outer fold, we center the training representations, compute an SVD, and retain the first 16 right singular vectors. Validation and test data use the training mean and basis. PCA is never fitted on validation or test episodes, and no whitening is applied.

\paragraph{Learned bottleneck.} The learned bottleneck has widths $256\rightarrow64\rightarrow r$, GELU activations, and $r\in\{16,8\}$. A 14-dimensional head predicts the training-only residual target from the current state and bottleneck output. AdamW uses learning rate $10^{-3}$, weight decay $10^{-5}$, batch size 128, and 100 epochs. We evaluate checkpoints at epochs 25, 50, 75, and 100 with seeds $\{101,202,303\}$. Validation selects settings lexicographically by GeoLASR, WorstLASR, parameter count, and configuration identifier. The reported summary directly aggregates $5$ folds $\times\,3$ seeds $\times\,2$ targets.

\paragraph{Ridge and width checks.} The ridge check reads cached representations aligned to the same transitions, so it performs no new model inference. Its fixed coefficient grid is $\{0,10^{-8},10^{-6},10^{-4},10^{-2}\}$. The grid is used only for sensitivity analysis, not for selecting a different main result. At the main coefficient $10^{-6}$, the cached reproduction differs from the main table by at most $3.8\times10^{-7}$ because of numerical rounding.

The width check maps each representation to 32 dimensions using PCA fitted on the outer training split. Representations that are already 32-dimensional remain unchanged. Table~\ref{tab:supp-sensitivity} reports both checks.

\begin{table}[!htbp]
\centering
\small
\setlength{\tabcolsep}{3pt}
\begin{tabular}{lccc}
\hline
\textbf{Representation} & \textbf{Main} & \textbf{Ridge sweep} & \textbf{32-D check} \\
\hline
DiLA & 0.750 & all $<1$ & $<1$ \\
LaWAM & 0.891 & all $<1$ & $<1$ \\
LAPA$^*$ & 0.922 & all $<1$ & $<1$ \\
UniVLA & 1.009 & all $\geq1$; $\dagger$ & $\geq1$; $\dagger$ \\
MVP-LAM & 1.095 & all $\geq1$ & $\geq1$; $\dagger$ \\
\hline
\end{tabular}
\caption{GeoLASR sensitivity relative to the boundary at one (lower is better). Main uses ridge $10^{-6}$; the sweep covers $\{0,10^{-8},10^{-6},10^{-4},10^{-2}\}$. The entries \emph{all $<1$} and \emph{all $\geq1$} mean that every sweep value lies on the stated side; 32-D reports the projected status. A dagger marks a finite score orders of magnitude above both one and the representation's main score, indicating numerical instability of the evaluator under that check. It is not a new LASR threshold; only boundary status is interpreted. Scores are computed without clipping, and no check selects the main setting.}
\label{tab:supp-sensitivity}
\end{table}

\FloatBarrier
\section{Adapter Architecture, Training, and Uncertainty}

The adapter is a small transformation applied to the frozen 256-dimensional DiLA latent. The three modes differ only in how the transformed output is used: residual adds a correction, replacement uses only the transformed output, and augmentation concatenates the raw and transformed vectors.

\paragraph{Low-rank transformation.} A rank-$r$ block is
\begin{equation}
 T_m(h)=W_{\mathrm{up}}^{(m)}\operatorname{GELU}\!\left(W_{\mathrm{down}}^{(m)}\operatorname{LN}(h)\right).
 \label{eq:supp-adapter-block}
\end{equation}
Layer normalization has width 256, $\epsilon=10^{-5}$, and learnable affine parameters. Both linear maps include biases. With depth $M$, the blocks compose as $T=T_M\circ\cdots\circ T_1$.

Residual mode implements the correction in the main paper as
\begin{equation}
 H=Z_{\mathrm{raw}}+\alpha T(Z_{\mathrm{raw}})
   =Z_{\mathrm{raw}}+\Delta_\theta(Z_{\mathrm{raw}}),
 \label{eq:supp-adapter-map}
\end{equation}
where the shared scalar starts at $\alpha_0=0$. Replacement uses $H=T(Z_{\mathrm{raw}})$. Augmentation uses $H=[Z_{\mathrm{raw}},T(Z_{\mathrm{raw}})]$. The identity references are $Z_{\mathrm{raw}}$ for the 256-dimensional modes and $[Z_{\mathrm{raw}},0]$ for augmentation.

\paragraph{LASR-inspired training objective.} For each target $Y_j$, we first fit a state predictor on training episodes and freeze it. The remaining target is
\begin{equation}
 r_j=Y_j-\widehat{\mathbb E}[Y_j\mid S].
 \label{eq:supp-adapter-residual}
\end{equation}
This residual removes the part already predicted from the current state. We standardize each residual coordinate with training-only statistics, and an affine head predicts it from $H$. Let $\mathcal L_j^{\mathrm{train}}(H)$ be the resulting MSE.

The surrogate compares this loss with fixed state and progress reference losses:
\begin{align}
 D_j&=\min\{\mathcal L^{\mathrm{train}}_{\mathrm{state},j},
                 \mathcal L^{\mathrm{train}}_{\mathrm{progress},j}\},\\
 \widetilde\rho_j(H)&=
 \frac{\mathcal L_j^{\mathrm{train}}(H)+\varepsilon_{\mathrm{opt}}}
      {D_j+\varepsilon_{\mathrm{opt}}}.
 \label{eq:supp-adapter-ratio}
\end{align}
Here, $\varepsilon_{\mathrm{opt}}=10^{-8}$ is used only for numerical stability during adapter training; reported LASR uses $\epsilon=0$.

For the action and state-change targets, the objective is
\begin{align}
 \mathcal L_{\mathrm{adapter}}
 &=\frac{1}{2}\sum_j\log\widetilde\rho_j(H) \notag\\
 &\quad+\beta\gamma\log\sum_j
 \exp\!\left(\frac{\log\widetilde\rho_j(H)}{\gamma}\right) \notag\\
 &\quad+\lambda_{\mathrm{id}}\,
 \mathbb E\!\left[\|H-H_{\mathrm{id}}\|_2^2\right],
 \label{eq:supp-adapter-objective}
\end{align}
with $\gamma=0.1$. The first term lowers the average log ratio, the second gives more weight to the worse target, and the last discourages large changes to the raw representation. This objective is used only for training. The selected output is evaluated from scratch with the held-out LASR metric.

\paragraph{Optimization and selection.} For every fold and seed, validation searches over ranks $\{16,32,64\}$, depths $\{1,2\}$, $\beta\in\{0,0.25,0.5\}$, and $\lambda_{\mathrm{id}}\in\{0,0.001,0.01\}$. This gives 54 candidates per mode. AdamW uses learning rate $10^{-3}$, weight decay $10^{-5}$, batch size 128, and 100 epochs. We check epochs 25, 50, 75, and 100. There is no scheduler, patience rule, or gradient clipping.

Validation orders candidates by GeoLASR, WorstLASR, parameter count, and configuration identifier. Selection is repeated for every fold and seed. Seeds $\{101,202,303\}$ control initialization, minibatch order, and pseudo-label generation. Training heads are discarded before held-out evaluation.

\paragraph{Training controls.} Identity leaves the raw representation unchanged. Ordinary MSE uses the same state-residual targets without the LASR relative-risk aggregation. Random supervision replaces the residual labels with independent standard-normal labels of the same shape. Shuffled supervision permutes the residual labels across training transitions. Neither random nor shuffled supervision uses validation or test outcomes.

The adapter summaries aggregate $5$ folds $\times\,3$ seeds $\times\,2$ targets, or 30 held-out target--fold--seed scores.

\begin{table}[!htbp]
\centering
\small
\setlength{\tabcolsep}{2pt}
\begin{tabular}{llrrr}
\hline
\textbf{Mode} & \textbf{Signal} & \textbf{Dim.} & \textbf{Geo} & \textbf{Worst} \\
\hline
None & Identity & 256 & 0.750 & 0.796 \\
Residual & LASR & 256 & \textbf{0.745} & \textbf{0.785} \\
Residual & MSE & 256 & 0.759 & 0.810 \\
Residual & Random & 256 & 0.762 & 0.815 \\
Residual & Shuffled & 256 & 0.761 & 0.834 \\
Replacement & LASR & 256 & 0.770 & 0.855 \\
Replacement & Random & 256 & 1.006 & 1.065 \\
Replacement & Shuffled & 256 & 1.004 & 1.047 \\
Augmentation & LASR & 512 & 0.749 & 0.834 \\
Augmentation & MSE & 512 & 0.772 & 0.820 \\
\hline
\end{tabular}
\caption{Full adapter ablation. LASR denotes the LASR-inspired surrogate. Lower is better. Random and shuffled residual adapters retain the raw-latent skip connection, so they test the training signal rather than a purely random representation.}
\label{tab:supp-adapter-ablation}
\end{table}

\paragraph{Paired uncertainty.} We resample whole episodes rather than individual transitions because transitions within one episode are related. The five test splits yield 49 episode-level vectors, one for each episode when it is held out. Each of 10,000 bootstrap replicates samples 49 episodes with replacement and computes a paired difference; the predictors are not refitted. For adapter comparisons, we first average the three training seeds for each fold--episode pair. The interval endpoints are the 2.5th and 97.5th percentiles. No additional test or multiple-comparison correction is applied.

\begin{table}[!htbp]
\centering
\small
\setlength{\tabcolsep}{3pt}
\begin{tabular}{lc}
\hline
\textbf{Comparator} & \textbf{$\Delta$GeoLASR [95\% CI]} \\
\hline
Raw latent & $-0.00563\;[-0.01852,\;0.00713]$ \\
Identity residual & $-0.00563\;[-0.01843,\;0.00745]$ \\
Random residual & $-0.01804\;[-0.02978,\;-0.00574]$ \\
Shuffled residual & $-0.01617\;[-0.02810,\;-0.00356]$ \\
MSE residual & $-0.01466\;[-0.02527,\;-0.00365]$ \\
\hline
\end{tabular}

\smallskip

\begin{tabular}{lc}
\hline
\textbf{Comparator} & \textbf{$\Delta$WorstLASR [95\% CI]} \\
\hline
Raw latent & $-0.00488\;[-0.01783,\;0.00899]$ \\
Identity residual & $-0.00488\;[-0.01797,\;0.00884]$ \\
Random residual & $-0.01713\;[-0.02956,\;-0.00439]$ \\
Shuffled residual & $-0.01456\;[-0.02718,\;-0.00109]$ \\
MSE residual & $-0.01404\;[-0.02533,\;-0.00284]$ \\
\hline
\end{tabular}
\caption{Paired episode-bootstrap differences: LASR-guided residual adapter minus comparator. Negative values favor the adapter. The paired estimates are not obtained by subtracting the separately aggregated rows in Table~\ref{tab:supp-adapter-ablation}.}
\label{tab:supp-adapter-bootstrap}
\end{table}

The intervals include zero when the residual adapter is compared with the raw latent or identity residual. They are below zero for the comparisons with random, shuffled, and ordinary-MSE residual training. Thus, the paired comparisons favor LASR training over the random, shuffled, and ordinary-MSE controls. The differences from the raw and identity representations remain point-estimate comparisons.

\FloatBarrier
\section{World-Model Prediction Details}

This experiment explores whether the representation change measured by LASR is also visible to the world-model predictor. We use the same LASR-guided residual adapter as in the main paper ablation. The adapter replaces the native DiLA latent with $H$, while the world-model predictor remains frozen.

The evaluation contains 128 Omni-Primitive-Transforms episodes. Original and adapted DiLA receive the same contexts and are compared with the same target frames. We report one-step and 15-step prediction. DINO uses the DINOv2 base checkpoint with register tokens~\citep{DBLP:journals/tmlr/OquabDMVSKFHMEA24}. Its exact identifier is \texttt{facebook/dinov2-with-registers-base}. SSIM uses an $11\times11$ Gaussian window with $\sigma=1.5$ and data range one~\citep{wang2004image}. LPIPS uses \texttt{lpips==0.1.4}, the VGG backbone, and inputs mapped from $[0,1]$ to $[-1,1]$~\citep{zhang2018unreasonable}.

\begin{table}[!htbp]
\centering
\small
\setlength{\tabcolsep}{2.5pt}
\begin{tabular}{llrrr}
\hline
\textbf{Model} & \textbf{Step} & \textbf{DINO$\downarrow$} & \textbf{SSIM$\uparrow$} & \textbf{LPIPS$\downarrow$} \\
\hline
Original DiLA & 1 & 0.389 & 0.818 & 0.302 \\
Adapted DiLA & 1 & \textbf{0.378} & \textbf{0.824} & \textbf{0.288} \\
Original DiLA & 15 & 0.521 & 0.828 & 0.339 \\
Adapted DiLA & 15 & \textbf{0.494} & \textbf{0.830} & \textbf{0.324} \\
\hline
\end{tabular}
\caption{World-model prediction with the LASR-guided residual adapter from the main paper ablation. The world-model predictor is frozen. Bold marks the better point estimate at each horizon.}
\label{tab:supp-world-transfer}
\end{table}

All six point estimates move in the preferred direction. We treat these values as descriptive evidence that the adapted representation also supports better prediction outside the LASR evaluator.

\FloatBarrier
\section{Extended Validation Suite}

This section reports additional validation experiments that complement the ALOHA
study. They use frozen public checkpoints and prespecified train/validation/test
separation. Unless otherwise stated, uncertainty uses 10,000 paired resamples
over fan or episode units. These experiments broaden the empirical contract;
they do not convert LASR into a causal or executability certificate.

\begin{table*}[!t]
\centering
\small
\setlength{\tabcolsep}{4pt}
\begin{tabular}{@{}lcccc@{}}
\toprule
\textbf{Route} & \textbf{GeoLASR [95\% CI]} &
\textbf{WorstLASR [95\% CI]} & \textbf{Support cells} &
\textbf{Branch acc. (\%)} \\
\midrule
DiLA native 16-frame & 0.688 [0.668,0.708] & 0.719 [0.697,0.771] & 10/10 & 65.58 \\
DiLA pair-only & 0.685 [0.665,0.704] & 0.714 [0.694,0.767] & 10/10 & -- \\
LAPA tokens & 0.831 [0.810,0.851] & 0.849 [0.831,0.908] & 10/10 & 50.39 \\
LaWAM & 0.860 [0.838,0.883] & 0.870 [0.859,0.933] & 10/10 & 48.59 \\
MVP-LAM & 0.988 [0.981,0.995] & 0.993 [0.988,1.013] & 10/10 & 26.75 \\
UniVLA controllable & 1.010 [0.998,1.025] & 1.049 [1.003,1.126] & 4/10 & 25.44 \\
\midrule
Recorded action$^\dagger$ & $\approx0$ & 0.002 [0.0018,0.0023] & 10/10 & -- \\
Matched random & 1.000 [0.998,1.003] & 1.001 [1.001,1.010] & 3/10 & -- \\
Within-fan shuffled & 0.999 [0.996,1.001] & 1.002 [0.999,1.009] & 8/10 & -- \\
Current-frame visual & 1.000 [1.000,1.000] & 1.000 [1.000,1.000] & 0/10 & -- \\
Next-frame outcome & 0.985 [0.981,0.990] & 0.994 [0.986,1.009] & 10/10 & -- \\
\bottomrule
\end{tabular}
\caption{Matched-panel results. Branch accuracy executes the common probe's
held-out decoded action from the exact saved source state; chance and the
state-only route are 25\%. The next-frame row is an outcome-leakage
counterexample rather than a valid executable-action positive.
$^\dagger$Unranked positive reference.}
\label{tab:supp-postfreeze-main}
\end{table*}

\subsection{Five-Task Matched Panel}

The panel contains PickCube-v1, PullCube-v1, PushCube-v1, StackCube-v1, and
PegInsertionSide-v1 from ManiSkill3~\citep{tao2024maniskill3}. We collect 103
exact-state fans per task and four action branches per fan, for 515 fans and
2,060 transitions. Every branch in a fan shares the same current simulator
state and current RGB frame. Targets are the recorded 7D action and measured
controllable TCP effect. The five public checkpoints were frozen before panel
generation, excluding exact episode overlap. Broader task-family and simulator
overlap remain possible.

All learned routes use the same transitions, targets, five fan-disjoint folds,
train-only PCA to eight dimensions, train-fitted target standardization, and
validation-only selection over ridge coefficients
$\{10^{-6},10^{-4},10^{-2},1,10^2,10^3\}$. The base control pool contains task
identity, fan progress, and a fixed current-frame RGB grid. LAPA's four
categorical token slots are one-hot encoded before train-only projection.

The pair-only DiLA route repeats the current frame 15 times and uses the next
frame as frame 16, exposing only the same $t,t+1$ information available to
pair-based models. Relative to native-16, its paired GeoLASR difference is
$-0.00313$ [$-0.00518,-0.00120$]; its WorstLASR difference is $-0.00409$
[$-0.00880,0.00592$]. Extra pre-action history therefore does not explain the
matched-panel result.

\begin{table}[!htbp]
\centering
\scriptsize
\setlength{\tabcolsep}{2.5pt}
\begin{tabular}{@{}lccccc@{}}
\toprule
\textbf{Task} & \textbf{DiLA} & \textbf{LAPA} & \textbf{LaWAM} &
\textbf{MVP} & \textbf{UniVLA} \\
\midrule
PickCube & .632 & .737 & .769 & .967 & 1.042 \\
PullCube & .596 & .735 & .734 & .970 & 1.008 \\
PushCube & .597 & .729 & .739 & .975 & 1.007 \\
StackCube & .607 & .749 & .763 & .994 & .998 \\
PegInsertionSide & 1.015 & 1.214 & 1.312 & 1.036 & .992 \\
\bottomrule
\end{tabular}
\caption{Task-stratified GeoLASR. The pooled result is not universal:
PegInsertionSide is a clear failure case for DiLA, LAPA, and LaWAM.}
\label{tab:supp-task-stratified}
\end{table}

\subsection{Matched Probe Families and Expanded Controls}

We apply the same route matching and split discipline to ridge, 64-feature
random Fourier features (RFF), and a small validation-selected MLP. The MLP
uses real optimizer/backward updates; all 110 target--fold--route fits changed
parameters and jointly executed 11,000 optimizer steps.

\begin{table}[!htbp]
\centering
\scriptsize
\setlength{\tabcolsep}{1.5pt}
\begin{tabular}{@{}lcccccc@{}}
\toprule
& \multicolumn{2}{c}{\textbf{Ridge}} &
\multicolumn{2}{c}{\textbf{RFF-64}} &
\multicolumn{2}{c}{\textbf{Small MLP}} \\
\cmidrule(lr){2-3}\cmidrule(lr){4-5}\cmidrule(lr){6-7}
\textbf{Route} & \textbf{Geo} & \textbf{Worst} &
\textbf{Geo} & \textbf{Worst} & \textbf{Geo} & \textbf{Worst} \\
\midrule
DiLA & .688 & .719 & .673 & .708 & .671 & .699 \\
LAPA & .831 & .849 & .793 & .815 & .763 & .783 \\
LaWAM & .860 & .870 & .830 & .859 & .794 & .818 \\
MVP-LAM & .988 & .993 & .989 & .995 & .991 & .998 \\
UniVLA & 1.010 & 1.049 & 196.969 & $2.86\!\times\!10^{11}$ & 1.002 & 1.011 \\
\bottomrule
\end{tabular}
\caption{Probe-family sensitivity. UniVLA's RFF result is a preserved numerical
instability, not evidence of an intrinsic model score.}
\label{tab:supp-probe-family}
\end{table}

The expanded control pool adds state squares, sine/cosine coordinates,
pairwise products, task-phase polynomials, scene-style statistics, and
spatial-layout pooling. It changes the selected control in 22 of 110 route
cells, corresponding to two target--fold decisions. In both, state-only remains the
stronger denominator, so route-level GeoLASR changes are below
$4.2\times10^{-10}$ and no admission changes. A focused synthetic check also
verifies that the matched polynomial-state control rejects a state-squared
recoding rather than treating it as new action information.

\subsection{Direct Contrast with Prediction Gain}

When the denominator is fixed, GeoLASR and negative raw prediction gain have
the same within-panel ordering. LASR adds a stronger admission rule: the latent
must also outperform the validation-selected control. In a matched-32D audit,
20/50 learned-route cells have positive state-only gain, but eight are rejected
by LASR. The representative LAPA fold-0 action cell has
$E_Z=2.3075\times10^{-4}<E_{\mathrm{state}}=2.3113\times10^{-4}$,
but $E_Z>E_{\mathrm{ctrl}}=2.2926\times10^{-4}$ and LASR is 1.0065.
In a separate six-dose action-corruption audit, 35/36 cells beat state-only,
yet 16 fail the stronger progress or temporal-shift control. LASR's added value
is therefore control-qualified admission, dimensionless cross-target
aggregation, and worst-cell auditing rather than a different fixed-denominator
rank order.

\subsection{One-Step Physical Replay}

Every public model's held-out common-probe action is executed from the exact
saved source state. Across 24,720 one-step replays, four-way branch accuracy is
65.58\% for DiLA, 50.39\% for LAPA, 48.59\% for LaWAM, 26.75\% for MVP-LAM,
and 25.44\% for UniVLA, versus 25\% state-only. Across the 25 model--task cells,
GeoLASR associates with executed TCP-effect MSE at task-block Spearman
$\rho=0.956$ [0.916,0.993] and branch error at $\rho=0.810$ [0.270,0.987].
This is held-out teacher-forced one-step execution under a common decoder; it
does not establish autonomous latent generation or native-policy control.

\subsection{Native-Policy Audit on the Same Distribution}

We run the official UniVLA policy on ten LIBERO-spatial tasks, ten episodes per
task and five seeds~\citep{liu2023libero}. All 500 episodes complete, yielding
55,162 aligned transitions and 477 successes. Folds group task and initial-state
index so the same registered initial state across seeds never crosses
train/validation/test. Targets are executed 7D action and 8D proprioceptive
state delta; every route is matched to width eight.

\begin{table}[!htbp]
\centering
\scriptsize
\setlength{\tabcolsep}{1.5pt}
\begin{tabular}{@{}lcc@{}}
\toprule
\textbf{Route} & \textbf{Geo [95\% CI]} &
\textbf{Worst [95\% CI]} \\
\midrule
Native hidden mean & .7747 [.7643,.7849] & .8418 [.8348,.8681] \\
Native token one-hot & 1.0402 [1.0302,1.0497] & 1.0765 [1.0635,1.1041] \\
Recorded action & $2.87\times10^{-11}$ & .3349 [.3212,.3549] \\
Matched random & 1.1270 [1.1161,1.1373] & 1.1819 [1.1659,1.2085] \\
Within-episode shuffle & 1.1210 [1.1102,1.1311] & 1.1738 [1.1577,1.2001] \\
Current-frame visual & 1.0093 [.9999,1.0183] & 1.0439 [1.0318,1.0730] \\
\bottomrule
\end{tabular}
\caption{UniVLA LASR on its official LIBERO-spatial rollout distribution.}
\label{tab:supp-univla-native}
\end{table}

The hidden route passes all ten cells while random and shuffled routes fail.
However, across 50 task--seed blocks, GeoLASR versus success has Spearman
$\rho=-0.1610$ with interval [$-0.6048,0.3930$]. The prespecified within-task
analysis gives $\rho=+0.2327$ with lower-tail stratified permutation
$p=0.8992$; leave-one-task-out pooled estimates range from $-0.3542$ to
$+0.0378$. This null prevents using LASR as a fine-grained policy-success
score. An official MVP-LAM evaluation separately completes 100/100 planned
episodes with 95 successes (Wilson 95\% interval [0.8882,0.9785]); because that
run lacks a matched LASR trace, it is evidence of a second runnable native
simulation path, not a cross-model LASR correlation.

\FloatBarrier
\subsection{Camera and Outcome Interventions}

Each intervention contains five tasks, 80 exact-state fans, four branches per
fan, and 320 paired transitions. Both conditions use the same frozen DiLA
route, five folds, two targets, width eight, expanded control pool, and 10,000
paired resamples.

\begin{table}[!htbp]
\centering
\scriptsize
\setlength{\tabcolsep}{2pt}
\begin{tabular}{@{}lcccc@{}}
\toprule
\textbf{Stress} & \textbf{Geo$_{0\rightarrow1}$} &
\textbf{$\Delta$Geo [95\% CI]} & \textbf{Worst$_{0\rightarrow1}$} &
\textbf{Flips} \\
\midrule
Camera & .7444$\rightarrow$.7625 & +.0181 [$-.0136,+.0501$] & .7769$\rightarrow$.8166 & 0/10 \\
Outcome & .7442$\rightarrow$.9888 & +.2446 [+.2051,+.2844] & .7739$\rightarrow$1.0018 & 2/10 \\
\bottomrule
\end{tabular}
\caption{Registered paired interventions. The camera condition changes the
rendering while holding the physical transition fixed. The outcome condition
holds history, current state/frame, action, and action-induced targets fixed,
then changes the realized post-transition object position before final
rendering.}
\label{tab:supp-interventions}
\end{table}

All 320 latents change under each intervention. Camera-pose admission is stable
in this registered shift, but the representation is not invariant. The outcome
intervention changes every latent and flips two cells, directly demonstrating
that the adjacent-frame route can track realized outcomes. We therefore require
outcome stress as a companion audit and do not interpret LASR alone as causal
action identification or executability.

\section{Computing and Reproducibility}

The representation and adapter experiments used two NVIDIA RTX 3090 24GB GPUs with Ubuntu 20.04.6 and Python 3.8.18. World-model prediction used one NVIDIA A100 80GB GPU. Its retained environment contains Python 3.11.15, PyTorch 2.6.0 with CUDA 12.4, NumPy 2.4.6, Pillow 12.3.0, and Transformers 5.13.0. The historical extraction environments varied across third-party models, and exact wall-clock times were not retained for every route.

The companion code artifact includes the LASR evaluator, panel construction and validation utilities, five extraction interfaces, representation diagnostics, adapter architecture and inference routes, frozen world-transfer evaluation, configuration files, and tests. It documents the public source, release or checkpoint name, extraction rules, and license for each third-party component. It does not include third-party data or repositories, model weights, latent caches, trained adapter weights, the adapter training loop, scalar experimental results, fold assignments, or paired-bootstrap inputs.

{\small
\bibliographystyle{ieeenat_fullname}
\bibliography{references}

@inproceedings{tao2024maniskill3,
  title={Maniskill3: Gpu parallelized robot simulation and rendering for generalizable embodied ai},
  author={Tao, Stone and Xiang, Fanbo and Shukla, Arth and Qin, Yuzhe and Hinrichsen, Xander and Yuan, Xiaodi and Bao, Chen and Lin, Xinsong and Liu, Yulin and Chan, Tse-Kai and others},
  booktitle={7th Robot Learning Workshop: Towards Robots with Human-Level Abilities},
  year={2025}
}

@inproceedings{liu2023libero,
  title={LIBERO: Benchmarking Knowledge Transfer for Lifelong Robot Learning},
  author={Liu, Bo and Zhu, Yifeng and Gao, Chongkai and Feng, Yihao and Liu, Qiang and Zhu, Yuke and Stone, Peter},
  booktitle={Advances in Neural Information Processing Systems},
  volume={36},
  pages={44776--44791},
  year={2023}
}

@inproceedings{zhang2024prelar,
  title={PreLAR: World Model Pre-training with Learnable Action Representation},
  author={Zhang, Lixuan and Kan, Meina and Shan, Shiguang and Chen, Xilin},
  booktitle={European Conference on Computer Vision},
  pages={185--201},
  year={2024},
  doi={10.1007/978-3-031-73337-6_11}
}

@article{khazatsky2024droid,
  title={Droid: A large-scale in-the-wild robot manipulation dataset},
  author={Khazatsky, Alexander and Pertsch, Karl and Nair, Suraj and Balakrishna, Ashwin and Dasari, Sudeep and Karamcheti, Siddharth and Nasiriany, Soroush and Srirama, Mohan Kumar and Chen, Lawrence Yunliang and Ellis, Kirsty and others},
  journal={arXiv preprint arXiv:2403.12945},
  year={2024}
}

@article{losey2022learning,
  title={Learning latent actions to control assistive robots},
  author={Losey, Dylan P and Jeon, Hong Jun and Li, Mengxi and Srinivasan, Krishnan and Mandlekar, Ajay and Garg, Animesh and Bohg, Jeannette and Sadigh, Dorsa},
  journal={Autonomous robots},
  volume={46},
  number={1},
  pages={115--147},
  year={2022},
  publisher={Springer}
}

@inproceedings{ye2025latent,
  title={Latent action pretraining from videos},
  author={Ye, Seonghyeon and Jang, Joel and Jeon, Byeongguk and Joo, Se June and Yang, Jianwei and Peng, Baolin and Mandlekar, Ajay and Tan, Reuben and Chao, Yu-Wei and Lin, Bill Yuchen and others},
  booktitle={International Conference on Learning Representations},
  volume={2025},
  pages={28213--28239},
  year={2025}
}

@inproceedings{bu2025univla,
  title={Learning to Act Anywhere with Task-centric Latent Actions},
  author={Bu, Qingwen and Yang, Yanting and Cai, Jisong and Gao, Shenyuan and Ren, Guanghui and Yao, Maoqing and Luo, Ping and Li, Hongyang},
  booktitle={Proceedings of Robotics: Science and Systems},
  year={2025},
  address={Los Angeles, CA, USA},
  month={June},
  doi={10.15607/RSS.2025.XXI.014}
}

@inproceedings{lee2026mvp,
  title={{MVP-LAM}: Learning Action-Centric Latent Action via Cross-Viewpoint Reconstruction},
  author={Lee, Jung Min and Lee, Dohyeok and Ju, Seokhun and Cho, Taehyun and Koo, Jin Woo and Zhao, Li and Hong, Sangwoo and Lee, Jungwoo},
  booktitle={Proceedings of the 43rd International Conference on Machine Learning},
  pages={64503--64524},
  year={2026},
  volume={306},
  series={Proceedings of Machine Learning Research},
  publisher={PMLR},
  url={https://proceedings.mlr.press/v306/lee26am.html}
}

@article{chen2026lawam,
  title={Lawam: Latent world action models for efficient dynamics-aware robot policies},
  author={Chen, Jialei and Wang, Kai and Chen, Kang and Chen, Shuaihang and Gao, Feng and Tang, Wenhao and Li, Zhiyuan and Liu, Weilin and Yao, Zhuyu and Li, Boxun and others},
  journal={arXiv preprint arXiv:2606.15768},
  year={2026}
}

@article{zhang2026dila,
  title={DiLA: Disentangled Latent Action World Models},
  author={Zhang, Tianqiu and Lyu, Muyang and Zhang, Yufan and Fang, Fang and Wu, Si},
  journal={arXiv preprint arXiv:2605.15725},
  year={2026}
}

@article{zhao2023learning,
  title={Learning fine-grained bimanual manipulation with low-cost hardware},
  author={Zhao, Tony Z and Kumar, Vikash and Levine, Sergey and Finn, Chelsea},
  journal={arXiv preprint arXiv:2304.13705},
  year={2023}
}

@inproceedings{hewitt2019designing,
  title={Designing and interpreting probes with control tasks},
  author={Hewitt, John and Liang, Percy},
  booktitle={Proceedings of the 2019 conference on empirical methods in natural language processing and the 9th international joint conference on natural language processing (emnlp-ijcnlp)},
  pages={2733--2743},
  year={2019}
}

@article{de2019causal,
  title={Causal confusion in imitation learning},
  author={De Haan, Pim and Jayaraman, Dinesh and Levine, Sergey},
  journal={Advances in neural information processing systems},
  volume={32},
  year={2019}
}

@article{alain2016understanding,
  title={Understanding intermediate layers using linear classifier probes},
  author={Alain, Guillaume and Bengio, Yoshua},
  journal={arXiv preprint arXiv:1610.01644},
  year={2016}
}

@article{wang2004image,
  title={Image quality assessment: from error visibility to structural similarity},
  author={Wang, Zhou and Bovik, Alan C and Sheikh, Hamid R and Simoncelli, Eero P},
  journal={IEEE transactions on image processing},
  volume={13},
  number={4},
  pages={600--612},
  year={2004},
  publisher={IEEE}
}

@inproceedings{zhang2018unreasonable,
  title={The unreasonable effectiveness of deep features as a perceptual metric},
  author={Zhang, Richard and Isola, Phillip and Efros, Alexei A and Shechtman, Eli and Wang, Oliver},
  booktitle={Proceedings of the IEEE conference on computer vision and pattern recognition},
  pages={586--595},
  year={2018}
}

@article{heusel2017gans,
  title={Gans trained by a two time-scale update rule converge to a local nash equilibrium},
  author={Heusel, Martin and Ramsauer, Hubert and Unterthiner, Thomas and Nessler, Bernhard and Hochreiter, Sepp},
  journal={Advances in neural information processing systems},
  volume={30},
  year={2017}
}

@inproceedings{zhang2019bertscore,
  title={{BERTScore}: Evaluating Text Generation with {BERT}},
  author={Zhang, Tianyi and Kishore, Varsha and Wu, Felix and Weinberger, Kilian Q and Artzi, Yoav},
  booktitle={International Conference on Learning Representations},
  year={2020},
  url={https://openreview.net/forum?id=SkeHuCVFDr}
}

@inproceedings{hessel2021clipscore,
  title={Clipscore: A reference-free evaluation metric for image captioning},
  author={Hessel, Jack and Holtzman, Ari and Forbes, Maxwell and Le Bras, Ronan and Choi, Yejin},
  booktitle={Proceedings of the 2021 conference on empirical methods in natural language processing},
  pages={7514--7528},
  year={2021}
}

@inproceedings{huang2024vbench,
  title={Vbench: Comprehensive benchmark suite for video generative models},
  author={Huang, Ziqi and He, Yinan and Yu, Jiashuo and Zhang, Fan and Si, Chenyang and Jiang, Yuming and Zhang, Yuanhan and Wu, Tianxing and Jin, Qingyang and Chanpaisit, Nattapol and others},
  booktitle={Proceedings of the IEEE/CVF Conference on Computer Vision and Pattern Recognition},
  pages={21807--21818},
  year={2024}
}

@article{ghosh2023geneval,
  title={Geneval: An object-focused framework for evaluating text-to-image alignment},
  author={Ghosh, Dhruba and Hajishirzi, Hannaneh and Schmidt, Ludwig},
  journal={Advances in Neural Information Processing Systems},
  volume={36},
  pages={52132--52152},
  year={2023}
}

@article{banerjee2005clustering,
  title={Clustering with Bregman divergences},
  author={Banerjee, Arindam and Merugu, Srujana and Dhillon, Inderjit S and Ghosh, Joydeep},
  journal={Journal of machine learning research},
  volume={6},
  number={Oct},
  pages={1705--1749},
  year={2005}
}

@article{blackwell1953equivalent,
  title={Equivalent comparisons of experiments},
  author={Blackwell, David},
  journal={The annals of mathematical statistics},
  pages={265--272},
  year={1953},
  publisher={JSTOR}
}

@article{gneiting2007strictly,
  title={Strictly proper scoring rules, prediction, and estimation},
  author={Gneiting, Tilmann and Raftery, Adrian E},
  journal={Journal of the American statistical Association},
  volume={102},
  number={477},
  pages={359--378},
  year={2007},
  publisher={Taylor \& Francis}
}

@inproceedings{gelada2019deepmdp,
  title={Deepmdp: Learning continuous latent space models for representation learning},
  author={Gelada, Carles and Kumar, Saurabh and Buckman, Jacob and Nachum, Ofir and Bellemare, Marc G},
  booktitle={International conference on machine learning},
  pages={2170--2179},
  year={2019},
  organization={PMLR}
}

@inproceedings{hafner2019learning,
  title={Learning latent dynamics for planning from pixels},
  author={Hafner, Danijar and Lillicrap, Timothy and Fischer, Ian and Villegas, Ruben and Ha, David and Lee, Honglak and Davidson, James},
  booktitle={International conference on machine learning},
  pages={2555--2565},
  year={2019},
  organization={PMLR}
}

@inproceedings{seo2022reinforcement,
  title={Reinforcement learning with action-free pre-training from videos},
  author={Seo, Younggyo and Lee, Kimin and James, Stephen L and Abbeel, Pieter},
  booktitle={International Conference on Machine Learning},
  pages={19561--19579},
  year={2022},
  organization={PMLR}
}

@inproceedings{schmidt2024learning,
  title={Learning to act without actions},
  author={Schmidt, Dominik and Jiang, Minqi},
  booktitle={International Conference on Learning Representations},
  volume={2024},
  pages={9379--9395},
  year={2024}
}

@inproceedings{bruce2024genie,
  title={Genie: Generative interactive environments},
  author={Bruce, Jake and Dennis, Michael D and Edwards, Ashley and Parker-Holder, Jack and Shi, Yuge and Hughes, Edward and Lai, Matthew and Mavalankar, Aditi and Steigerwald, Richie and Apps, Chris and others},
  booktitle={Forty-first International Conference on Machine Learning},
  year={2024}
}

@article{gao2025adaworld,
  title={Adaworld: Learning adaptable world models with latent actions},
  author={Gao, Shenyuan and Zhou, Siyuan and Du, Yilun and Zhang, Jun and Gan, Chuang},
  journal={arXiv preprint arXiv:2503.18938},
  year={2025}
}

@article{liang2025clam,
  title={Clam: Continuous latent action models for robot learning from unlabeled demonstrations},
  author={Liang, Anthony and Czempin, Pavel and Hong, Matthew and Zhou, Yutai and Biyik, Erdem and Tu, Stephen},
  journal={arXiv preprint arXiv:2505.04999},
  year={2025}
}

@inproceedings{zhang2026latent,
  author = {Zhang, Chuheng and Pearce, Tim and Zhang, Pushi and Wang, Kaixin and Chen, Xiaoyu and Shen, Wei and Zhao, Li and Bian, Jiang},
 booktitle = {Advances in Neural Information Processing Systems},
 editor = {D. Belgrave and C. Zhang and H. Lin and R. Pascanu and P. Koniusz and M. Ghassemi and N. Chen},
 pages = {146676--146697},
 publisher = {Curran Associates, Inc.},
 title = {What Do Latent Action Models Actually Learn?},
 url = {https://proceedings.neurips.cc/paper_files/paper/2025/file/d7b3cef7c31b94a4a533db83d01a8882-Paper-Conference.pdf},
 volume = {38},
 year = {2025}
}

@inproceedings{garrido2026learning,
  title={Learning Latent Action World Models In The Wild},
  author={Garrido, Quentin and Nagarajan, Tushar and Terver, Basile and Ballas, Nicolas and LeCun, Yann and Rabbat, Michael},
  booktitle={Proceedings of the 43rd International Conference on Machine Learning},
  pages={34113--34150},
  year={2026},
  volume={306},
  series={Proceedings of Machine Learning Research},
  publisher={PMLR},
  url={https://proceedings.mlr.press/v306/garrido26a.html}
}

@article{jiang2026olaf,
  title={Olaf-World: Orienting Latent Actions for Video World Modeling},
  author={Jiang, Yuxin and Gu, Yuchao and Tsang, Ivor W and Shou, Mike Zheng},
  journal={arXiv preprint arXiv:2602.10104},
  year={2026}
}

@inproceedings{nikulin2025latent,
  title={Latent Action Learning Requires Supervision in the Presence of Distractors},
  author={Nikulin, Alexander and Zisman, Ilya and Tarasov, Denis and Lyubaykin, Nikita and Polubarov, Andrei and Kiselev, Igor and Kurenkov, Vladislav},
  booktitle={Proceedings of the 42nd International Conference on Machine Learning},
  pages={46427--46447},
  volume={267},
  series={Proceedings of Machine Learning Research},
  publisher={PMLR},
  year={2025}
}

@article{wei2026causally,
  title={Causally Debiased Latent Action Model for Embodied Action Conditioned World Models},
  author={Wei, Yufan and Zhou, Kun and Mao, Lingjun and Zhang, Zijun and Xu, Ziming and Xi, Ziqiao and Liang, Shuang and Han, Ruobing and Yan, Yuchen and Wang, Xinyue and others},
  journal={arXiv preprint arXiv:2607.09185},
  year={2026}
}

@article{lee2026latent,
  title={Why Latent Actions Fail, and How to Prevent It},
  author={Lee, Jung Min and Cho, Taehyun and Zhao, Li and Lee, Jungwoo},
  journal={arXiv preprint arXiv:2605.20223},
  year={2026}
}

@inproceedings{hewitt2021conditional,
  title={Conditional probing: measuring usable information beyond a baseline},
  author={Hewitt, John and Ethayarajh, Kawin and Liang, Percy and Manning, Christopher D},
  booktitle={Proceedings of the 2021 Conference on Empirical Methods in Natural Language Processing},
  pages={1626--1639},
  year={2021}
}

@inproceedings{qin2024worldsimbench,
  title={WorldSimBench: Towards Video Generation Models as World Simulators},
  author={Qin, Yiran and Shi, Zhelun and Yu, Jiwen and Wang, Xijun and Zhou, Enshen and Li, Lijun and Yin, Zhenfei and Liu, Xihui and Sheng, Lu and Shao, Jing and Bai, Lei and Zhang, Ruimao},
  booktitle={Proceedings of the 42nd International Conference on Machine Learning},
  pages={50338--50362},
  volume={267},
  series={Proceedings of Machine Learning Research},
  publisher={PMLR},
  year={2025}
}

@article{lachapelle2025identifiability,
  title={On the Identifiability of Latent Action Policies},
  author={Lachapelle, S{\'e}bastien},
  journal={arXiv preprint arXiv:2510.01337},
  year={2025}
}

@article{lin2026pixels,
  title={From Pixels to Tokens: A Systematic Study of Latent Action Supervision for Vision-Language-Action Models},
  author={Lin, Yihan and Li, Haoyang and Li, Yang and Shen, Haitao and Zhao, Yihan and Shao, Chao and Zhang, Jing},
  journal={arXiv preprint arXiv:2605.04678},
  year={2026}
}

@article{cui2024dynamo,
  title={Dynamo: In-domain dynamics pretraining for visuo-motor control},
  author={Cui, Zichen J and Pan, Hengkai and Iyer, Aadhithya and Haldar, Siddhant and Pinto, Lerrel},
  journal={Advances in Neural Information Processing Systems},
  volume={37},
  pages={33933--33961},
  year={2024}
}

@inproceedings{chen2025moto,
  title={Moto: Latent motion token as the bridging language for learning robot manipulation from videos},
  author={Chen, Yi and Ge, Yuying and Tang, Weiliang and Li, Yizhuo and Ge, Yixiao and Ding, Mingyu and Shan, Ying and Liu, Xihui},
  booktitle={Proceedings of the IEEE/CVF International Conference on Computer Vision},
  pages={19752--19763},
  year={2025}
}

@article{chen2025villa,
  title={Villa-x: enhancing latent action modeling in vision-language-action models},
  author={Chen, Xiaoyu and Wei, Hangxing and Zhang, Pushi and Zhang, Chuheng and Wang, Kaixin and Guo, Yanjiang and Yang, Rushuai and Wang, Yucen and Xiao, Xinquan and Zhao, Li and others},
  journal={arXiv preprint arXiv:2507.23682},
  year={2025}
}

@inproceedings{klepach2026object,
  title={Object-centric latent action learning},
  author={Klepach, Albina and Nikulin, Alexander and Zisman, Ilya and Tarasov, Denis and Derevyagin, Alexander and Polubarov, Andrei and Lyubaykin, Nikita and Kiselev, Igor and Kurenkov, Vladislav},
  booktitle={Proceedings of the AAAI Conference on Artificial Intelligence},
  volume={40},
  pages={22626--22634},
  year={2026}
}

@inproceedings{bu2026laof,
  title={Laof: Robust latent action learning with optical flow constraints},
  author={Bu, Xizhou and Lyu, Jiexi and Sun, Fulei and Yang, Ruichen and Ma, Zhiqiang and Li, Wei},
  booktitle={Proceedings of the IEEE/CVF Conference on Computer Vision and Pattern Recognition},
  pages={27334--27344},
  year={2026}
}

@article{vafa2024evaluating,
  title={Evaluating the world model implicit in a generative model},
  author={Vafa, Keyon and Chen, Justin Y and Rambachan, Ashesh and Kleinberg, Jon and Mullainathan, Sendhil},
  journal={Advances in Neural Information Processing Systems},
  volume={37},
  pages={26941--26975},
  year={2024}
}

@misc{jha2026reconstruction,
      title={Reconstruction or Semantics? What Makes a Latent Space Useful for Robotic World Models}, 
      author={Nilaksh and Saurav Jha and Artem Zholus and Sarath Chandar},
      year={2026},
      eprint={2605.06388},
      archivePrefix={arXiv},
      primaryClass={cs.CV},
      url={https://arxiv.org/abs/2605.06388}, 
}

@article{zhang2025world,
  title={World-in-world: World models in a closed-loop world},
  author={Zhang, Jiahan and Jiang, Muqing and Dai, Nanru and Lu, Taiming and Uzunoglu, Arda and Zhang, Shunchi and Wei, Yana and Wang, Jiahao and Patel, Vishal M and Liang, Paul Pu and others},
  journal={arXiv preprint arXiv:2510.18135},
  year={2025}
}

@article{saanum2024simplifying,
  title={Simplifying latent dynamics with softly state-invariant world models},
  author={Saanum, Tankred and Dayan, Peter and Schulz, Eric},
  journal={Advances in Neural Information Processing Systems},
  volume={37},
  pages={38355--38382},
  year={2024}
}

@article{DBLP:journals/tmlr/OquabDMVSKFHMEA24,
  author       = {Maxime Oquab and
                  Timoth{\'{e}}e Darcet and
                  Th{\'{e}}o Moutakanni and
                  Huy V. Vo and
                  Marc Szafraniec and
                  Vasil Khalidov and
                  Pierre Fernandez and
                  Daniel Haziza and
                  Francisco Massa and
                  Alaaeldin El{-}Nouby and
                  Mido Assran and
                  Nicolas Ballas and
                  Wojciech Galuba and
                  Russell Howes and
                  Po{-}Yao Huang and
                  Shang{-}Wen Li and
                  Ishan Misra and
                  Michael Rabbat and
                  Vasu Sharma and
                  Gabriel Synnaeve and
                  Hu Xu and
                  Herv{\'{e}} J{\'{e}}gou and
                  Julien Mairal and
                  Patrick Labatut and
                  Armand Joulin and
                  Piotr Bojanowski},
  title        = {DINOv2: Learning Robust Visual Features without Supervision},
  journal      = {Trans. Mach. Learn. Res.},
  volume       = {2024},
  year         = {2024},
  url          = {https://openreview.net/forum?id=a68SUt6zFt},
  bibsource    = {dblp computer science bibliography, https://dblp.org}
}
}

\end{document}